\documentclass{article}

\usepackage{microtype}
\usepackage{graphicx}
\usepackage{tcolorbox}
\usepackage{booktabs} 

\usepackage{hyperref}

\usepackage[accepted]{icml2023}

\usepackage{amsmath}
\usepackage{amssymb}
\usepackage{mathtools}
\usepackage{amsthm}

\usepackage[capitalize,noabbrev]{cleveref}

\theoremstyle{plain}
\newtheorem{theorem}{Theorem}[section]
\newtheorem{proposition}[theorem]{Proposition}
\newtheorem{lemma}[theorem]{Lemma}
\newtheorem{corollary}[theorem]{Corollary}
\theoremstyle{definition}

\theoremstyle{remark}

\newcommand{\mcal}[1]{\mathcal{#1}}

\newcommand{\risk}[1]{R(#1)}

\newcommand{\cipsrisk}[2]{\hat{R}^{#2}_n(#1)}
\newcommand{\cvcipsrisk}[3]{\hat{R}^{#2, #3}_n(#1)}

\DeclareMathOperator*{\argmin}{arg\,min}
\DeclareMathOperator*{\argmax}{arg\,max}

\usepackage{subcaption}
\usepackage{amsthm}
\usepackage{thm-restate}
\usepackage{optidef}
\usepackage{tabularx}
\usepackage{fixltx2e}
\usepackage{booktabs}
\usepackage{hhline}
\usepackage{makecell}
\usepackage{bbm}
\usepackage{graphicx}
\usepackage{tikz}
\usetikzlibrary{bayesnet}
\usetikzlibrary{arrows}
\usepackage{float}
\usepackage{bigints}

\usepackage[disable,textsize=tiny]{todonotes}

\icmltitlerunning{PAC-Bayesian Offline Contextual Bandits With Guarantees}

\begin{document}

\twocolumn[
\icmltitle{PAC-Bayesian Offline Contextual Bandits With Guarantees}



\icmlsetsymbol{equal}{*}

\begin{icmlauthorlist}
\icmlauthor{Otmane Sakhi}{ccc,eee}
\icmlauthor{Pierre Alquier}{sss}
\icmlauthor{Nicolas Chopin}{eee}
\end{icmlauthorlist}

\icmlaffiliation{ccc}{Criteo AI Lab, Paris, France}
\icmlaffiliation{eee}{CREST, ENSAE, IPP, Palaiseau, France}
\icmlaffiliation{sss}{ESSEC Business School, Asia-Pacific campus, Singapore}

\icmlcorrespondingauthor{Otmane Sakhi}{o.sakhi@criteo.com}

\icmlkeywords{Machine Learning, ICML}

\vskip 0.3in
]



\printAffiliationsAndNotice{}  
\begin{abstract}

This paper introduces a new principled approach for off-policy learning in contextual bandits. Unlike previous work, our approach does not derive learning principles from intractable or loose bounds. We analyse the problem through the PAC-Bayesian lens, interpreting policies as mixtures of decision rules. This allows us to propose novel generalization bounds and provide tractable algorithms to optimize them. We prove that the derived bounds are tighter than their competitors, and can be optimized directly to confidently improve upon the logging policy \textit{offline}. Our approach learns policies with guarantees, uses all available data and does not require tuning additional hyperparameters on held-out sets. We demonstrate through extensive experiments the effectiveness of our approach in providing performance guarantees in practical scenarios.
\end{abstract}

\section{Introduction}

Online industrial systems encounter sequential decision problems as they interact with the environment and strive to improve based on the received feedback. The contextual bandit framework formalizes this mechanism, and proved valuable with applications in recommender systems \cite{valko2014spectral} and clinical trials \cite{villar2015multi}. It describes a game of repeated interactions between a system and an environment, where the latter reveals a context that the system interacts with, and receives a feedback in return.
   
While the online solution to this problem involves strategies that find an optimal trade-off between exploration and exploitation to minimize the \textit{regret} \cite{tor}, we are concerned with its offline formulation \cite{swaminathan2015batch}, which is arguably better suited for real-life applications, where more control over the decision-maker, often coined \textit{the policy}, is needed. The learning of the policy is performed offline, based on historical data, typically obtained by logging the interactions between an older version of the decision system and the environment. By leveraging this data, our goal is to discover new strategies of greater performance.

There are two main paths to address this learning problem. The \textit{direct method} \cite{blob, pess} attacks the problem by modelling the feedback and deriving a policy according to this model. This approach can be praised for its simplicity \cite{overparam}, is well-studied in the offline setting \cite{dr-pessimism} but will often suffer from a bias as the feedback received is complex and the efficiency of the method directly depends on our ability to understand the problem's structure. 

We will consider the second path of off-policy learning, or IPS: \textit{inverse propensity scoring} \cite{ips} where we learn the policy directly from the logged data after correcting its bias with importance sampling \cite{is}. As these estimators \cite{bottou2015counterfactual, snips} can suffer from a variance problem as we drift away from the logging policy, the literature gave birth to different learning principles \cite{swaminathan2015batch, imitation, london2020bayesian, faury20distributionally} motivating penalizations toward the logging policy. These principles are inspired by generalization bounds, but introduce a hyperparameter $\lambda$ to either replace intractable quantities \cite{swaminathan2015batch} or to tighten a potentially vacuous bound \cite{london2020bayesian}. These approaches require tuning $\lambda$ on a held-out set and sometimes fail at improving the previous decision system \cite{london2020bayesian, surrogate}.

 In this work, we analyse off-policy learning from the PAC-Bayesian perspective \cite{mcallester, CatoniBound}. We aim at introducing a novel, theoretically-grounded approach, based on the direct optimization of newly derived tight generalization bounds, to obtain guaranteed improvement of the previous system \textit{\textbf{offline}}, without the need for held-out sets nor hyperparameter tuning. We show that our approach is perfectly suited to this framework, as it naturally incorporates information about the old decision system and can confidently improve it.

\section{Preliminaries}

\subsection{Setting}
    
    We use $x\in\mcal{X}$ to denote a context and $a\in \mathcal{A} = [K]$ an action, where $K$ denotes the number of available actions. For a context $x$, each action is associated with a cost $c(x,a)\in[-1,0]$, with the convention that better actions have smaller cost. The cost function $c$ is unknown. Our decision system is represented by its policy $\pi : \mathcal{X} \to \mathcal{P}(\mathcal{A})$ which given $x\in\mcal{X}$, defines a probability distribution over the discrete action space $\mathcal{A}$ of size $K$. Assuming that the contexts are stochastic and follow an unknown distribution $\nu$, we define the \emph{risk} of the policy $\pi$ as the expected cost one suffers when playing actions according to $\pi$:
    \begin{align*}
        \risk{\pi} = \mathbbm{E}_{x\sim\nu, a\sim\pi(\cdot|x)}\left[c(x,a)\right].
    \end{align*}
    The learning problem is to find a policy $\pi$ which minimizes the risk. This risk can be naively estimated by deploying the policy online and gathering enough interactions to construct an accurate estimate. Unfortunately, we do not have this luxury in most real-world problems as the cost of deploying bad policies can be extremely high. We can obtain instead an estimate by exploiting the logged interactions collected by the previous system. Indeed, the previous system is represented by a \emph{logging policy} $\pi_0$ (e.g a previous version of a recommender system that we are trying to improve), which gathered interaction data of the following form:
    \begin{align*}
        \mcal{D}_n= \left\{x_i,a_i\sim\pi_0(\cdot|x_i), c_i\right\}_{i\in[n]},\quad\mbox{with } c_i =c(x_i, a_i).
    \end{align*}
    Given this data, one can build various estimators, with the clipped IPS \cite{bottou2015counterfactual} the most commonly used. It is constructed based on a clipping of the importance weights or the logging propensities to mitigate variance issues \cite{cips}. We are more interested in clipping the logging probabilities as we need objectives that are linear in the policy $\pi$ for our study. The cIPS estimator is given by:
    \begin{align}
    \label{cIPS}
        \cipsrisk{\pi}{\tau} = \frac{1}{n}\sum_{i=1}^n \frac{\pi(a_i|x_i)}{\max \{\pi_0(a_i|x_i), \tau\}}c_i
    \end{align}
    
with $\tau \in [0, 1]$ being the clipping factor. 
     
     Choosing $\tau \ll 1$ reduces the bias of cIPS. We recover the classical IPS estimator \cite{ips} (unbiased under mild conditions) by taking $\tau = 0$.
     
     Another estimator with better statistical properties is the doubly robust estimator \cite{ben2013robust}, which uses the importance weights as control variates to reduce further the variance of the cIPS estimators. This estimator is asymptotically optimal \cite{more_dr} (in terms of variance) amongst the class of unbiased and consistent off-policy estimators.
     
     We consider a simplified version of this estimator, which replaces the use of a model $\hat{c}$ of the cost by one parameter $\xi \in [-1, 0]$ that can be chosen freely. We define the control variate clipped IPS, or cvcIPS as follows:
    \begin{align}
    \label{cvcIPS}
        \cvcipsrisk{\pi}{\tau}{\xi} = \xi + \frac{1}{n}\sum_{i=1}^n \frac{\pi(a_i|x_i)}{\max \{\pi_0(a_i|x_i), \tau\}}(c_i - \xi).
    \end{align}
    The cvcIPS estimator can be seen as a special case of the doubly robust estimator when the cost model $\hat{c} = \xi$ is constant and $\tau = 0$. cIPS is recovered by setting $\xi = 0$. This simple estimator is deeply connected to the \textit{SNIPS} estimator \cite{snips} and was shown to be more suited to off-policy learning as it mitigates the problem of propensity overfitting \cite{deep}.

\subsection{Related Work: Learning Principles}

The literature so far has focused on deriving new principles to learn policies with good online performance. The first line of work in this direction is CRM: Counterfactual Risk minimization \cite{swaminathan2015batch} which adopted SVP: Sample Variance Penalization \cite{emp_bern} to favor policies with small empirical risk and controlled variance. The intuition behind it is that the variance of cIPS depends on the disparity between $\pi$ and $\pi_0$ making the estimator unreliable when $\pi$ drifts away from $\pi_0$. The analysis focused on the cIPS estimator and used uniform bounds based on empirical Bernstein inequalities \cite{emp_bern}, where intractable quantities were replaced by a tuning parameter $\lambda$, giving the following learning objective:
\begin{align}
    \label{SVP}
        \argmin_\pi \left\{ \cipsrisk{\pi}{\tau} + \lambda \sqrt{\frac{\hat{V}_n(\pi)}{n}} \right\}
\end{align}
with  $\hat{V}_n(\pi)$ the empirical variance of the cIPS estimator. A majorization-minimization algorithm was provided in \cite{swaminathan2015batch} to solve Equation \eqref{SVP} for parametrized softmax policies.

In the same spirit, \cite{faury20distributionally, dro_2} generalize SVP using the distributional robustness framework, showing that the CRM principle can be retrieved with a particular choice of the divergence and provide asymptotic coverage results of the true risk. Their objectives are competitive with SVP while providing simple ways to scale its optimization to large datasets.

Another line of research, closer to our work, uses PAC-Bayesian bounds to derive learning objectives in the same fashion as \cite{swaminathan2015batch}. Indeed, \cite{london2020bayesian} introduce the Bayesian CRM, motivating the use of $L_2$ regularization towards the parameter $\theta_0$ of the logging policy $\pi_0$. The analysis uses \cite{mcallester2}'s bound, is conducted on the cIPS estimator and controls the $L_2$ norm by a hyperparameter $\lambda$, giving the following learning objective for parametrized softmax policies:
\begin{align}
    \label{PB0}
        \argmin_\theta \left\{ \cipsrisk{\pi_\theta}{\tau} + \lambda ||\theta - \theta_0||^2 \right\}.
\end{align}
\cite{london2020bayesian} minimize a convex upper-bound of objective \eqref{PB0} (by taking a log transform of the policy) which is amenable to stochastic optimization, giving better results than \eqref{SVP} while scaling better to the size of the dataset.

\paragraph{Limitations.} The principles found in the literature are mainly inspired by generalization bounds. However, the bounds from where these principles are derived either depend on intractable quantities \cite{swaminathan2015batch} or are not tight enough to be used as-is \cite{london2020bayesian}. For example, \cite{swaminathan2015batch} derive a generalisation bound (see Theorem 1 in \cite{swaminathan2015batch}) for offline policy learning using the notion of covering number. This introduces the complexity measure $\mathcal{Q}_\mcal{H}(n, \gamma)$
 that cannot be computed (even for simple policy classes) making their bound intractable. This forces the introduction of a hyperparameter $\lambda$ that needs further tuning. Unfortunately, this approach suffers from numerous problems:
\begin{itemize}
    \item \textbf{No Theoretical Guarantees.} Introducing the hyperparameter $\lambda$ in Equations \eqref{SVP} and \eqref{PB0} gives tractable objectives, but loses the theoretical guarantees given by the initial bounds. These objectives do not necessarily cover the true risk, and optimizing them can lead to  policies worse than the logging $\pi_0$. Empirical evidence can be found in \cite{surrogate, london2020bayesian} where the SVP principle in Equation \eqref{SVP} fails to improve on $\pi_0$.

    \item \textbf{Inconsistent Strategy.} These principles were first introduced to mitigate the suboptimality of learning with off-policy estimators, deemed untrustworthy for their potential high variance. The strategy minimizes the objectives for different values of $\{\lambda_1, ..., \lambda_m\}$, generating a set of policy candidates $\{\pi_{\lambda_1}, ..., \pi_{\lambda_m}\}$, from which we select the best policy $\pi_{\lambda_*}$ according to the same untrustworthy, high variance estimators on a held-out set. This makes the selection strategy used inconsistent with what these principles are claiming to solve.

        \item \textbf{Tuning requires additional care.} Tuning $\lambda$ needs to be done on a held-out set. This means that we need to train multiple policies (computational burden) on a fraction of the data (data inefficiency), and select the best policy among the candidates using off-policy estimators (variance problem) on the held-out set.
        
\end{itemize}
In this paper, we derive a coherent principle, that learns policies using all the available data and provides guarantees on their performance, without the introduction of new hyperparameters.

\subsection{Learning With Guaranteed Improvements}
\label{sec:gi}

Our first concern in most applications is to improve upon the actual system $\pi_0$. As $\mcal{D}_n$ is collected by $\pi_0$, we  have access to $\risk{\pi_0}$\footnote{up to a small $\mcal{O}(1/\sqrt{n})$ approximation error.}. Given a new policy $\pi$, we want to be confident that the improvement $\mathcal{I}(\pi, \pi_0) = \risk{\pi_0} - \risk{\pi}$ is positive before deployment. 

Let us suppose that we are restricted to a class of policies $\mcal{H}$, and have access to a generalization bound that gives the following result with high probability over draws of $\mcal{D}_n$:
$$\risk{\pi} \le \mathcal{UB}_n(\pi) \quad \forall \pi \in \mcal{H}.$$
with $\mathcal{UB}_n$ an empirical upper bound that depends on $\mcal{D}_n$. 
For any $\pi$, we define the guaranteed  improvement: 
$$\mcal{GI}_{\mcal{UB}_n}(\pi, \pi_0) = \risk{\pi_0} - \mathcal{UB}_n(\pi).$$ 
We can be sure of improving $\risk{\pi_0}$ offline if we manage to find $\pi \in \mcal{H}$ that achieves $\mcal{GI}_{\mcal{UB}_n}(\pi, \pi_0) > 0$ as the following result will hold with high probability: 
\begin{align*}
    \mathcal{I}(\pi, \pi_0) \ge \mcal{GI}_{\mcal{UB}_n}(\pi, \pi_0) > 0.
\end{align*}
To obtain such a policy, we look for the minimizer of $\mathcal{UB}_n$ over the class of policies $\mathcal{H}$ as:
\begin{align*}
\pi^*_{\mathcal{UB}_n} \in \argmin_{\pi \in \mathcal{H}} \mathcal{UB}_n(\pi) = \argmax_{\pi \in \mathcal{H}} \mcal{GI}_{\mcal{UB}_n}(\pi, \pi_0).
\end{align*}
We define the best guaranteed risk and the best guaranteed improvement follows:
\begin{align*}
\mcal{GR}^*_{\mcal{UB}_n} &= \mcal{UB}_n(\pi^*_{\mcal{UB}_n})\\
\mcal{GI}^*_{\mcal{UB}_n}(\pi_0) &= \risk{\pi_0} - \mcal{GR}^*_{\mcal{UB}_n}.
\end{align*}

\begin{tcolorbox}
A \textit{theoretically-grounded} strategy to improve $\pi_0$ will be to deploy $\pi^*_{\mcal{UB}_n}$ if we obtain a positive guaranteed improvement $\mcal{GI}^*_{\mcal{UB}_n}(\pi_0) > 0$, otherwise continue collecting data with the current system $\pi_0$.
\end{tcolorbox}

This strategy will always produce policies that are at least as good as $\pi_0$, optimizes directly a bound over all data and does not require held-out sets nor new hyperparameters. However, the tightness of the bounds $\mcal{UB}_n$ will play an important role. Indeed, If we fix $\mcal{D}_n$ and $\pi_0$, $\mcal{GI}^*_{\mcal{UB}_n}(\pi_0)$ will only depend on the minimum of $\mcal{UB}_n$, motivating the derivation of bounds that are tractable and tight enough to achieve the smallest minimum possible. 

In this regard, we opt for the PAC-Bayesian framework to tackle this problem as it is proven to give tractable, non-vacuous bounds even in difficult settings \cite{dl}. Its paradigm also fits our application as we can incorporate information about the previous system $\pi_0$ in the form of a prior; see \cite{bayes_alquier} for a recent review. 

\paragraph{Contributions.} We advocate for a theoretically grounded strategy that uses generalization bound to improve $\pi_0$ with guarantees. So far, the existing bounds are either intractable \cite{swaminathan2015batch} or can be proven to be suboptimal \cite{london2020bayesian}. In this work, 
\begin{itemize}
    \item we derive new, tractable and tight generalization bounds using the PAC-Bayesian framework. These bounds are fully tractable unlike \cite{swaminathan2015batch}'s bound and are tighter than \cite{london2020bayesian}'s bound.
    \item we provide a way to optimize our bounds over a particular class of policies and show empirically that they can guarantee improvement over $\pi_0$ in practical scenarios.
\end{itemize}

\section{Motivating PAC-Bayesian tools}

As previously discussed, in the contextual bandit setting, we seek a policy that minimizes the expected cost:  
\begin{align*}
        \risk{\pi} = \mathbbm{E}_{x\sim\nu, a\sim\pi(\cdot|x)}\left[c(x,a)\right].
    \end{align*}
The minimizer of this objective over the unrestricted space of policies is a deterministic decision rule defined by:
\begin{align*}
        \forall x, a \quad \pi^*(a|x) = \mathbbm{1}[\operatorname*{argmin}_{a'} c(x,a') = a].
    \end{align*}
Given a context $x$, the solution will always choose the action that has the minimum cost. However, as the function $c$ is generally unknown, we instead learn a parametric score function $f_\theta \in \mcal{F}_\Theta = \{f_\theta:\mathcal{X} \times [K] \to \mathbbm{R}, \theta \in \Theta\}$ that encodes the action's relevance to a context $x$. Given a function $f_\theta$, we define the decision rule $d_\theta$ by:
$$d_\theta(a|x) = \mathbbm{1}[\operatorname*{argmax}_{a'} f_\theta(x,a') = a].$$ 
These parametric decisions rules will be the building blocks of our analysis. We view stochastic policies as smoothed decision rules, with smoothness induced by distributions over the space of score functions $\mcal{F}_\Theta$. Given a context $x$, instead of sampling an action $a$ directly from a distribution on the action set, we sample a function $f_\theta$ from a distribution over $\mcal{F}_\Theta$ and compute the action as $a = \operatorname*{argmax}_{a'} f_\theta(x, a')$. With this interpretation, for any
distribution $Q$ on $\mcal{F}_\Theta$, the probability of an action, $a \in \mathcal{A}$, given a context $x \in \mathcal{X}$, is defined as the expected value of $d_\theta$ over $Q$, that is:
\begin{align*}
    \pi_Q(a|x) &= \mathbbm{E}_{{\theta \sim Q}}\left[d_\theta(a|x)\right]\\
    &= \mathbbm{P}_{Q}\left(\operatorname*{argmax}_{a'} f_\theta(x,a') = a\right).
\end{align*}

\paragraph{Policies as mixtures  of decision rules.} This perspective on policies was introduced in \cite{pbps} and later developed in \cite{london2020bayesian}. Constructing policies as \textit{mixtures of deterministic decision rules} does not restrict the class of policies our study applies to. Indeed, if the family $\mcal{F}_\Theta$ is rich enough (e.g, neural networks), we give the following theorem that proves that any policy $\pi$ can be written as a mixture of deterministic policies.

\begin{theorem}
\label{existence}
Let us fix a policy $\pi$. Then there is a probability distribution $Q_\pi$ on the set of all functions $f:\mathcal{X}\times[K] \rightarrow \{0,1\}$ such that
$$
\forall x,a \quad \pi(a|x) = \mathbbm{E}_{f \sim Q_\pi}\left[\mathbbm{1}\left[\operatorname*{argmax}_{a'} f(x,a') = a\right]\right].
$$
\end{theorem}

A formal proof of Theorem \ref{existence} is given in Appendix~\ref{mixpol}. This means that adopting this perspective on policies does not narrow the scope of our study. For a policy $\pi_Q$ defined by a distribution $Q$ over $\mcal{F}_\Theta$, we observe that by linearity, its true risk can be written as:
$$\risk{\pi_Q} = \mathbbm{E}_{\theta \sim Q}[\risk{d_\theta}].$$
Similarly, clipping the logging propensities in our empirical estimators allows us to obtain a linear estimators in $\pi$. For instance, we can estimate empirically the risk of the policy $\pi_Q$ with cvcIPS (as it generalizes cIPS) and obtain:
\begin{equation*}
    \cvcipsrisk{\pi_Q}{\tau}{\xi} = \mathbbm{E}_{\theta \sim Q}[\cvcipsrisk{d_\theta}{\tau}{\xi}].
\end{equation*}
By linearity, one can see that both the true and empirical risk of a policy $\pi_Q$ can also be interpreted as the average risk of decision rules drawn from the distribution $Q$. This duality is in the heart of our analysis and paves the way nicely to the PAC-Bayesian framework, which studies generalization properties of the average risk of randomized predictors \cite{bayes_alquier}. If we fix a reference distribution $P$ over $\mcal{F}_\Theta$ and define the KL divergence from $P$ to $Q$ as:
\begin{align*}
    \mcal{KL}(Q||P) = 
    \begin{cases}
    \bigintss  \ln \left\{\frac{dQ}{dP} \right\} dQ \text{~ if $Q$ is $P$-continuous,}\\
    +\infty \text{~ otherwise,}
    \end{cases}
\end{align*}
we can construct with the help of PAC-Bayesian tools bounds holding for the average risk of decision rules over any distribution $Q$;
$$\mathbbm{E}_{\theta \sim Q}[\risk{d_\theta}] \le \mathbbm{E}_{\theta \sim Q}[\cvcipsrisk{d_\theta}{\tau}{\xi}] + \mathcal{O}\left(\mathcal{KL}(Q||P)\right).$$
Our objective will be to find tight generalisation bounds of this form as this construction, coupled with the linearity of our objective and estimator, allows us to obtain tight bounds holding for any policy $\pi_Q$;
$$\risk{\pi_Q} \le \cvcipsrisk{\pi_Q}{\tau}{\xi} + \mathcal{O}\left(\mathcal{KL}(Q||P)\right).$$

\textbf{The PAC-Bayesian Paradigm.} Before we dive deeper into the analysis, we want to emphasize the similarities between the PAC-Bayesian paradigm and the offline contextual bandit problem. This learning framework proceeds as follows: Given a class of functions $\mcal{F}_\Theta$, we fix a prior (reference distribution) $P$ on $\mcal{F}_\Theta$ before seeing the data, then, we receive some data $\mcal{D}_n$ which help us learn a better distribution $Q$ over $\mcal{F}_\Theta$ than our reference $P$. With the previous perspective on policies, the prior $P$, even if it can be any data-free distribution, will be our logging policy (i.e. $\pi_0 = \pi_P$), and we will use the data $\mcal{D}_n$ to learn distribution $Q$, thus a new policy $\pi_Q$ that improves the logging policy $\pi_0$. 

\section{PAC-Bayesian Analysis}
\subsection{Bounds for clipped IPS}

The clipped IPS estimator \cite{bottou2015counterfactual} is often studied for offline policy learning \cite{swaminathan2015batch, london2020bayesian} as it is easy to analyse, and have a negative bias (once the cost is negative) facilitating the derivation of learning bounds.

\cite{london2020bayesian} adapted \cite{mcallester2}'s bound to derive their learning objective. We state a slightly tighter version in Proposition \ref{mcall} for the cIPS estimator. The proof of this bound cannot be adapted naively to the cvcIPS estimator, because once $\xi \neq 0$, the bias of the estimator, an intractable quantity that depends on the unknown distribution $\nu$, is no longer negative and needs to be incorporated in the bound, making the bound itself intractable.

\begin{proposition} \label{mcall} Given a prior $P$ on $\mcal{F}_\Theta$, $\tau \in (0, 1]$, $\delta \in (0,1]$, the following bound holds with probability at least $1 - \delta$ uniformly for all distribution $Q$ over $\mcal{F}_\Theta$:
\begin{multline*}
    \risk{\pi_Q} \le \cipsrisk{\pi_Q}{\tau} + \frac{2(\mcal{KL}(Q||P) + \ln\frac{2\sqrt{n}}{\delta})}{\tau n} \\ + \sqrt{\frac{2[\cipsrisk{\pi_Q}{\tau} + \frac{1}{\tau}](\mcal{KL}(Q||P) + \ln\frac{2\sqrt{n}}{\delta})}{\tau n}}.
\end{multline*}

\end{proposition}

The upper bound stated in the previous proposition will be denoted by $\mcal{LS}^{P, \delta, \tau}_n(\pi_Q)$. When there is no ambiguity, we will also drop ${P, \delta, \tau}$ (all fixed) and only use $\mcal{LS}_n(\pi_Q)$. 

\cite{mcallester2}'s bound can give tight results in the $[0,1]$-bounded loss case when the empirical risk is close to 0 as one obtains fast convergence rates in $O(1/n)$. However, its use in the case of offline contextual bandits is far from being optimal. Indeed, to achieve a fast rate in this context, one needs $\cipsrisk{\pi_Q}{\tau} + \frac{1}{\tau} \approx 0$. This is hardly achievable in practice especially when $n$ is large and $\tau \ll 1$.

To defend our claim, let us suppose that for each context $x$, there is one optimal action $a_x^*$ for which $c(x, a_x^*) = -1$ and it is 0 otherwise. Let us write down the clipped IPS:
    \begin{align*}
        \cipsrisk{\pi}{\tau} = \frac{1}{n}\sum_{i=1}^n \frac{\pi(a_i|x_i)}{\max \{\pi_0(a_i|x_i), \tau\}}c_i \ge -\frac{1}{\tau}.
    \end{align*}
To get equality, we need:
$$\forall i \in [n], \quad c_i = -1, \quad \pi_0(a_i|x_i) \le \tau, \quad \pi(a_i|x_i) = 1.$$
If $n$ is large, the first condition on the costs means that $\pi_0$ is near optimal and the played actions $a_i$ are optimal. For this, we get that $\forall i \in [n], \pi_0(a_i|x_i) \approx 1$. This, combined with the second condition on $\pi_0$ gives that $\tau \approx 1$. In practice, $\pi_0$ is never the optimal policy and $\tau \ll 1$ which makes the fast rate condition $\cipsrisk{\pi}{\tau} + \frac{1}{\tau} \approx 0$ unachievable. In the majority of scenarios, as we penalize $\pi_Q$ to stay close to $\pi_0$ through the KL divergence, we will have $\cipsrisk{\pi_Q}{\tau} \in [-1, 0]$, thus $\cipsrisk{\pi_Q}{\tau} + \frac{1}{\tau} \approx \frac{1}{\tau}$, giving a limiting behavior:
\vspace{-0.5em}
$$\mcal{LS}_n(\pi_Q) = \cipsrisk{\pi_Q}{\tau} + \mathcal{O}\left(\frac{1}{\tau}\sqrt{\frac{\mathcal{KL}(Q||P)}{n}}\right).$$
If we want to get tighter results, we need to look for bounds with better dependencies on $\tau$ and $n$. In our pursuit of a tighter bound, we derive the following result:
\begin{proposition} \label{PBP_catoni} Given a prior $P$ on $\mcal{F}_\Theta$, $\tau \in (0,1]$, $\delta \in (0,1]$. The following bound holds with probability at least $1 - \delta$ uniformly for all distribution $Q$ over $\mcal{F}_\Theta$:

\resizebox{\hsize}{!}{
\begin{minipage}{\linewidth}
\begin{align*}
        \risk{\pi_Q} \le \min\limits_{\lambda > 0} \frac{1 - \exp\left( - \tau \lambda \cipsrisk{\pi_Q}{\tau} - \frac{1}{n}\left[\mcal{KL}(Q||P) + \ln \frac{2\sqrt{n}}{\delta} \right] \right)}{\tau(e^{\lambda} - 1)}
\end{align*}
\end{minipage}
}

\end{proposition}

We will denote by $\mcal{C}^{P, \delta, \tau}_n(\pi_Q)$ the upper bound stated in this proposition. When there is no ambiguity, we will also drop ${P, \delta, \tau}$ (all fixed) and only use $\mcal{C}_n(\pi_Q)$.

This is a direct application of \cite{CatoniBound}'s bound  to the bounded loss cIPS while exploiting the fact that its bias is negative. A full derivation can be found in Appendix~\ref{catoni_pbp}. Note that Proposition \ref{PBP_catoni} cannot be applied to the cvcIPS estimator ($\xi \neq 0$) as its bias is non-negative and intractable. To be able to measure the tightness of this bound, we estimate its limiting behaviour to understand its dependency on $\tau$ and $n$. We derive in Appendix~\ref{catoni_lb} the following result:
$$\mcal{C}_n(\pi_Q) =  \cipsrisk{\pi_Q}{\tau} + \mathcal{O}\left(\frac{\mathcal{KL}(Q||P)}{\tau n}\right).$$

This shows that $\mcal{C}_n$ has a better dependency on $n$ compared to $\mcal{LS}_n$. Actually, we can prove that $\mcal{C}_n$ will always give tighter results as the next theorem states that it is smaller than $\mcal{LS}_n$ in all scenarios.

\begin{theorem} \label{dominates}

For any $\mcal{D}_n \sim (\mu, \pi_0)^n$, any distributions $P, Q$, any $\tau \in (0, 1], \delta \in (0,1]$, we have:
$$\mcal{C}^{P, \delta, \tau}_n(\pi_Q) \le \mcal{LS}^{P, \delta, \tau}_n(\pi_Q).$$
\end{theorem}
One can refer to Appendix~\ref{thm1} for the full proof. This result confirms that the bound given by Proposition \ref{PBP_catoni} is \textit{theoretically} tighter than the bound in Proposition \ref{mcall}, making $\mcal{LS}_n$ unusable if we seek tight guarantees on $\risk{\pi_Q}$.

\subsection{Going beyond clipped IPS} \label{bern}

The cvcIPS estimator in Equation \eqref{cvcIPS} generalizes cIPS, and can behave better as it achieves improved variance with a well chosen $\xi$. To study its learning properties, we derive a \textbf{novel} \textit{Bernstein-type} PAC-Bayesian bound that holds for the cvcIPS estimator. Let $g$ be the function $g : u \xrightarrow[]{}\frac{\exp(u) - 1 - u}{u^2}$. We also define the conditional bias $\mcal{B}^{\tau}_{n}$ and the conditional second moment $\mathcal{V}_{n}^\tau$:
{\centering
\resizebox{\hsize}{!}{
\begin{minipage}{\linewidth}
\begin{align*}
\bullet \quad \mcal{B}^{\tau}_{n}(\pi) &= \frac{1}{n}\sum \limits_{i=1}^{n} \mathbbm{E}_{\pi(.|x_i)}\left[\mathbbm{1}[\pi_0(a|x_i) < \tau]  \left( 1 - \frac{\pi_0(a|x_i)}{\tau} \right)\right] \\
 \bullet \quad  \mathcal{V}_{n}^\tau(\pi) &= \frac{1}{n}\sum \limits_{i=1}^{n} \mathbbm{E}_{\pi(.|x_i)} \left[\frac{\pi_0(a|x_i)}{\max \{\pi_0(a|x_i), \tau\}^2}\right].
\end{align*}
\end{minipage}
}
}

With these definitions, we can state our proposition:
\begin{proposition} \label{pbp_tract_bernstein_cv} Given a prior $P$ on $\mcal{F}_\Theta$, $\xi \in [-1, 0], \tau \in (0, 1]$, $\delta \in (0,1]$ and a set of strictly positive scalars $\Lambda = \{ \lambda_i \}_{i \in [n_\Lambda]}$. The following bound holds with probability at least $1 - \delta$ uniformly for all distribution $Q$ over $\mcal{F}_\Theta$:
\vspace{-1em}

\resizebox{\hsize}{!}{
\begin{minipage}{\linewidth}
\begin{align*}
  \risk{\pi_Q} &\le \cvcipsrisk{\pi_Q}{\tau}{\xi} - \xi \mathcal{B}^{\tau}_{n}(\pi_Q) + \sqrt{\frac{\mcal{KL}(Q||P) + \ln \frac{4\sqrt{n}}{\delta}}{2 n}} \\
  &+ \min\limits_{\lambda \in \Lambda} \left\{\frac{\mcal{KL}(Q||P) + \ln \frac{2 n_\Lambda}{\delta}}{\lambda n} +  \lambda l_\xi g\left(\lambda b_\xi \right) \mathcal{V}_{n}^\tau(\pi_Q)\right\}
\end{align*}
\end{minipage}
}

with $l_{\xi} = \max \left[\xi^2, (1 + \xi)^2 \right]$, $b_\xi = (1 + \xi)/\tau - \xi$.
\end{proposition}
The choice of $\Lambda$ as well as the full proof of a more general version of Proposition \ref{pbp_tract_bernstein_cv} can be found in Appendix~\ref{proof_prop3}. The upper bound given by Proposition \ref{pbp_tract_bernstein_cv} (with $P, \tau, \delta$ fixed) will be denoted by $\mcal{CBB}^{\xi}_{n}(\pi_Q)$.

Proposition \ref{pbp_tract_bernstein_cv} covers the general cvcIPS estimator ($\xi \neq 0$) and its objective is decomposable into a sum, making it amenable to stochastic first order optimization \cite{rm_sgd}. However, minimizing it requires access to the logging policy $\pi_0$. This is reasonable as $\pi_0$ represents the currently deployed decision system, which we want to improve. In the limit, the bound is estimated to behave like:
\begin{align*}
    \mcal{CBB}^{\xi}_{n}(\pi_Q) &= \cvcipsrisk{\pi_Q}{\tau}{\xi} - \xi \mathcal{B}^{\tau}_{n}(\pi_Q)\\ 
    &+ \mathcal{O}\left( \left(\frac{1}{2\sqrt{2}} + \sqrt{l_\xi \mathcal{V}_{n}^\tau(\pi_Q)}\right)\sqrt{ \frac{\mathcal{KL}(Q||P)}{n}}\right)
\end{align*}
A derivation of this result can be found in Appendix~\ref{bigo}. We have $\mcal{V}^\tau_{n} \le 1/\tau$ and expect this bound to give tight results when $\mcal{V}^\tau_{n}(\pi_Q) \ll 1/\tau$. However, once $\pi_0$ is uniform and $\tau = 1/K$, we can never have the previous condition as: $$\forall \pi, \quad \mcal{V}^\tau_{n}(\pi) = 1/\tau.$$ 
This means that the worst regime for $\mcal{CBB}^{\xi}_{n}$ is when $\pi_0$ is uniform, and even in that case, this bound should be tighter than $\mcal{LS}_n$ as it has a better dependency on $\tau$. We can also get an intuition about the impact of $\xi$, in particular:
\begin{itemize}
    \item $\xi = 0$ recovers cIPS. This estimator has the best dependency on the bias (it nullifies the impact of $\mathcal{B}^{\tau}_{n}(\pi_Q)$) and the worst dependency on $\mcal{V}^\tau_{n}(\pi_Q)$  as $l_\xi = 1$.
    \item $\xi = - 0.5$ obtains the best dependency on $\mcal{V}^\tau_{n}(\pi_Q)$  as $l_\xi$ reaches its minimum and make the bound suffer only half the bias $\mathcal{B}^{\tau}_{n}(\pi_Q)$.
    \item $\xi = -1$ in the other hand, has both the worst dependencies on the bias and the variance, and can be considered a bad choice. Actually, this value of $\xi$ shifts the costs to be always positive ($\forall i, c_i - \xi \ge 0$), which is known to make the off-policy risk estimators not suited for policy learning \cite{swaminathan2015batch}.
\end{itemize}

These observations point to the importance of the choice of $\xi$, which can drastically change the behavior of the bound. We will empirically study the impact of two candidate values of $\xi \in \{0, -0.5\}$ on the tightness of the bound.

\section{Restricting the Space of Policies}

Our PAC-Bayesian bounds hold for any policy $\pi_Q$. However, to obtain policies of practical use, we ask for some desired properties that are summarized in the points below.

\textbf{Sampling.} Being able to efficiently sample actions from our policy is crucial as the decisions taken by our online system boil down to sampling. For a given context $x$, we have:
$$a \sim \pi_Q(\cdot|x) \iff a = \operatorname*{argmax}_{a'} f_\theta(x, a') \text{, } \theta \sim Q.$$
The complexity of sampling from $\pi_Q$ depends on how easy it is to sample from the distribution $Q$. 

\textbf{Computing propensities.} Computing propensities for a given pair $(x, a)$ is essential for off-policy evaluation. A generic estimate can be obtained by:
\begin{align*}
    \hat{\pi}^{\text{naive}}_Q(a|x) = \frac{1}{S}\sum_{i = 1}^S d_{\theta_i}(a|x)
\end{align*}
with $\{ \theta_i \}_{i = 1}^S$ samples from $Q$. This estimator will behave badly once we deal with large action spaces and/or high-dimensional distributions parameters. Ideally, we would like to exploit the family of distributions $Q$ considered and the form of the function $f_\theta$ to come up with a better behaved estimator for the propensities.

\textbf{Numerical optimization.} If we restrict our study to a parametric family $\mcal{Q}_\Psi = \{Q_\psi, \psi \in \Psi \}$, computing gradients will be essential to minimising the bounds. For a given pair $(x, a)$, we can compute for any parameterized distribution $Q_\psi$ the score function gradient estimator \cite{REINFORCE} of $\pi_{Q_\psi}(a|x)$:
\begin{align*}
    \nabla_\psi \pi_{Q_\psi}(a|x) &= \nabla_\psi \mathbbm{E}_{\theta \sim Q_\psi}[d_\theta(a|x)] \\
    &= \mathbbm{E}_{\theta \sim Q_\psi}[d_\theta(a|x) \nabla_\psi \log Q_\psi(\theta)].
\end{align*}
This gradient suffers from a variance problem \cite{reparam} and we might need to choose a specific family of distributions $\mcal{Q}_\Psi$, or/and specify a form of $f_\theta$ to obtain $\psi \rightarrow \pi_{Q_\psi}(a|x)$ with better behaved gradients.

\cite{london2020bayesian} restricted their study to Mixed Logit policies \cite{mixed_logits}. These policies are easy to sample from, and have easy to compute propensities and gradients. However, their learning properties are demonstrated to be sub-optimal \cite{soft_pull}. To this end, we adopt another class of policies that we deem better behaved for our objective. We discuss the reasons behind this choice in detail in Appendix~\ref{mixedl}.

\subsection{Linear Independent Gaussian Policies}

As mentioned previously, even if our analysis is valid for all distributions $Q$ and any form of $f_\theta$, we need to restrict our space to obtain practical policies. 
We restrict $f_\theta$ to:
\begin{align}
\label{scr_func}
    \forall x,a \quad f_\theta(x, a) = \phi(x)^T\theta_a
\end{align}
with $\phi$ a fixed transform\footnote{Even if we assume that $\phi$ is fixed, the analysis can be naturally extended to the more general case where $\phi_\psi$ is a parameterized neural network that we learn.} over the contexts. This form of $f_\theta$ is widely used in this context \cite{faury20distributionally,swaminathan2015batch}. This results in a parameter $\theta$ of dimension $d = p \times K$ with $p$ the dimension of the features $\phi(x)$ and $K$ the number of actions.

We also restrict the family of distributions $\mcal{Q}_{d + 1} = \{Q_{\boldsymbol{\mu}, \sigma} = \mathcal{N}(\boldsymbol{\mu}, \sigma^2 I_d), \boldsymbol{\mu} \in \mathbbm{R}^d, \sigma > 0\}$ to independent Gaussians with shared scale. 

With these choices of $f_\theta$ and $\mcal{Q}$, the induced $\pi_{\boldsymbol{\mu}, \sigma}$, that we call \textbf{LIG: Linear Independent Gaussian} policies, will provide fast sampling and easily computable propensities and gradients. Indeed, sampling from $\pi_{\boldsymbol{\mu}, \sigma}$ will reduce to sampling from a normal distribution $\theta \sim Q_{\boldsymbol{\mu}, \sigma}$ and computing $a = \operatorname*{argmax}_{a'} f_\theta(x, a')$. When it comes to estimating the propensity of $a$ given $x$, we can suggest another expression of $\pi_{\boldsymbol{\mu}, \sigma}(a|x)$ that reduces the computation to a one dimensional integral:
\resizebox{\hsize}{!}{
\begin{minipage}{\linewidth}
\begin{align*}
    \pi_{\boldsymbol{\mu}, \sigma}(a|x) &= \mathbbm{E}_{\epsilon \sim \mathcal{N}(0, 1)}\left[\prod_{a' \neq a} \Phi\left(\epsilon + \frac{\phi(x)^T(\boldsymbol{\mu}_a - \boldsymbol{\mu}_{a'})}{\sigma ||\phi(x)||}\right) \right] \\
    &= \mathbbm{E}_{\epsilon \sim \mathcal{N}(0, 1)}\left[ G_{\boldsymbol{\mu}, \sigma}(\epsilon, a, x)  \right]
\end{align*}
\end{minipage}
}

with $\Phi$ the cumulative distribution function of the standard normal. See Appendix~\ref{ligp} for a full derivation. The computation of $\pi_{\boldsymbol{\mu}, \sigma}(a|x)$ becomes easier as one dimensional standard normal integrals can be well approximated. The gradient can also be derived from this new expression:
$$\nabla_{\boldsymbol{\mu}, \sigma} \pi_{\boldsymbol{\mu}, \sigma}(a|x) = \mathbbm{E}_{\epsilon \sim \mathcal{N}(0, 1)}\left[ \nabla_{\mu, \sigma} G_{\mu, \sigma}(\epsilon, a, x)\right]
$$ 
which can be seen as a one dimensional reparametrization trick gradient, and is known to behave better than the score function gradient estimator \cite{reparam}.

\textbf{Optimising the bounds.} For their practicality, we focus on the class of \textbf{LIG} policies to optimise the bounds. As these policies are built with Gaussian distributions $Q$, we also adopt Gaussian priors\footnote{The prior uses the parameters of the logging policy $\pi_0$.} $P = \mcal{N}(\boldsymbol{\mu_0}, \sigma_0 I_d)$ to obtain an analytical expression for $\mcal{KL}(Q||P)$. We state the bounds for \textbf{LIG} policies with Gaussian priors in Appendix \ref{lig_bounds}.

Optimizing for \textbf{LIG} policies, the best guaranteed risk defined in Section \ref{sec:gi} and the minimizer $\pi^*_{\mcal{UB}_n}$ for the different bounds $\mathcal{UB}_n \in \{\mcal{LS}_n, \mcal{C}_n, \mcal{CBB}^\xi_n\}$ are given by:
\begin{align}
\label{gr}
\mcal{GR}^*_{\mcal{UB}_n} &= \min_{\pi_{\boldsymbol{\mu}, \sigma}}\mcal{UB}_n({\pi_{\boldsymbol{\mu}, \sigma}})\\
\pi^*_{\mcal{UB}_n} &= \argmin_{\pi_{\boldsymbol{\mu}, \sigma}} \mcal{UB}_n(\pi_{\boldsymbol{\mu}, \sigma}).
\end{align}
The best guaranteed improvement $\mcal{GI}^*(\pi_0)$ with these bounds follows as the difference between $R(\pi_0)$ and $\mcal{GR}^*$.

\begin{figure*}
     \centering
\includegraphics[width=1.\textwidth]{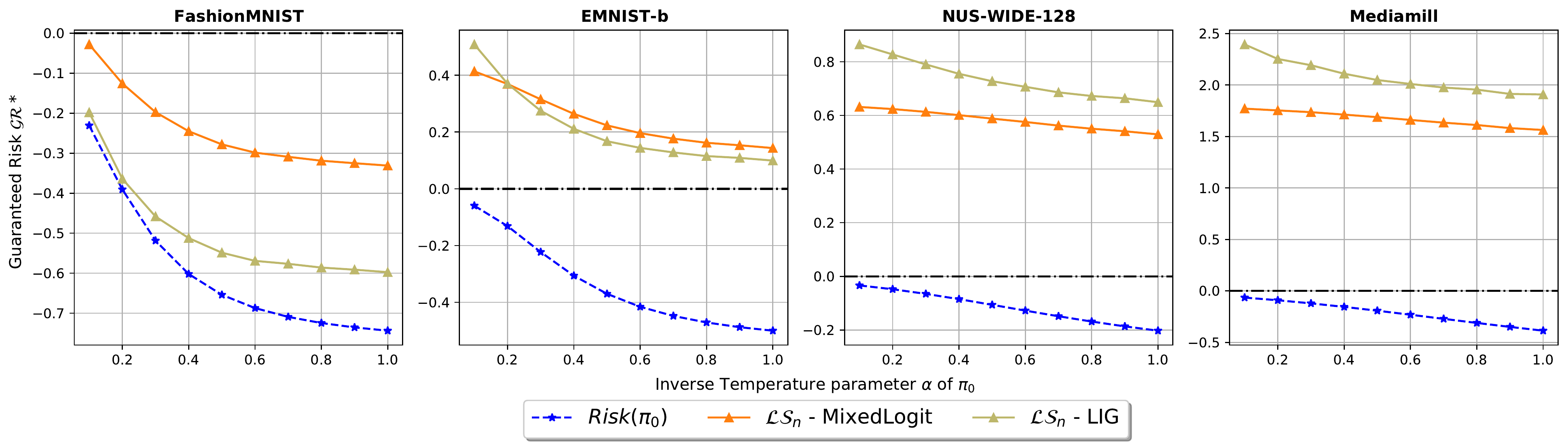}
    \caption{The Guaranteed Risk given by $\mcal{LS}_n$ optimized over \textbf{Mixed Logit} and \textbf{LIG} policy classes while changing $\pi_0$. The $\mcal{LS}_n$ bound fails to guarantee improvement ($\mcal{GR}^*_{\mcal{LS}_n} > \risk{\pi_0} $) in all scenarios considered.}
    \label{fig:ls}
\end{figure*}

\section{Experiments}

We are interested in studying the effectiveness of the proposed bounds in providing guaranteed improvement of the previously deployed system $\pi_0$ after collecting $\mcal{D}_n$. 

\textbf{Experimental Setup.} For ease of exposition, we use a softmax logging policy of parameter $\mu_0 \in \mathbbm{R}^{p \times K}$:
$$\pi^{\mcal{S}}_{\boldsymbol{\mu_0}}(a|x) \propto \exp(\phi(x)^T\boldsymbol{\mu_0}_a).$$ 
$\mu_{{0}_a} \in \mathbbm{R}^{p}$ is the parameter associated with action $a$. The policy $\pi^{\mcal{S}}_{\boldsymbol{\mu_0}}$ is used to generate the logged interactions data $\mcal{D}_n$ and its parameter $\boldsymbol{\mu_0}$ constructs the reference distribution $P = \mathcal{N}(\boldsymbol{\mu_0}, I_d)$ for all bounds. 

We adopt the standard supervised-to-bandit
conversion to generate logged data in all of our experiments \cite{swaminathan2015batch}. We use two mutliclass datasets: \textbf{FashionMNIST} \cite{fashionmnist} and \textbf{EMNIST-b} \cite{EMNIST-b}, alongside two multilabel datasets: \textbf{NUS-WIDE-128} \cite{nuswide} with 128-VLAD features \cite{vlad} and \textbf{Mediamill} \cite{mediamill} to empirically validate our findings. The statistics of the datasets are described in Table \ref{table:det_stats} in Appendix~\ref{detailed_exps}; $N$ the size of the training split, $K$ the number of actions and $p$ the dimension of the features $\phi(x)$. We take a small fraction ($5\%$) of the training data that will only be used to learn $\boldsymbol{\mu_0}$ in a supervised manner. 

With $\boldsymbol{\mu_0}$ obtained, we introduce an inverse temperature parameter $\alpha$ to our softmax logging policy $\pi^\mcal{S}_{\alpha \boldsymbol{\mu_0}}$ giving a prior $P = \mcal{N}(\alpha \boldsymbol{\mu_0}, I_d)$. Changing $\alpha$ allows us to cover logging policies with different entropies ($\alpha \approx 0$ gives a uniform $\pi_0$ and $\alpha \approx 1$ gives a peaked $\pi_0$). We run $\pi_{\alpha \boldsymbol{\mu_0}}$ on the rest ($n_c = 0.95N$) of the training data to generate $\mcal{D}_n$. For a context $x$ and an action $a$, we define in our setting the cost as $c = - \mathbbm{1}[a \in y]$ with $y$ the set of true labels for $x$.


Learning $\boldsymbol{\mu_0}$ on a split different than the one logged allows us to use the previous bounds as the parameter $\boldsymbol{\mu_0}$ does not depend on the logged interactions, making the reference distribution $P$ data-free.

We set the allowed uncertainty to $\delta = 0.05$. For all datasets, $\tau$ will be set to $\tau = 1/K$ to get no bias when the logging policy is close to uniform.  We use Adam \cite{adam} with a learning rate of $10^{-3}$ for 100 epochs to optimize the bounds w.r.t their parameters. More details on the training procedure can be found in Appendix~\ref{training}.

\textbf{$\mcal{LS}_n$ is not tight enough.} We demonstrated in this work that the $\mcal{LS}_n$ bound used by \cite{london2020bayesian} is suboptimal as it generally shows a worse dependence on $\tau$ and can be shown to be theoretically dominated by $\mcal{C}_n$ (Theorem~\ref{dominates}). However, we want to verify if it can guarantee the improvement of the logging policy $\pi_0$. Given logged data $\mcal{D}_n$ generated by $\pi_0$, we optimize the $\mcal{LS}_n$ bound with both the \textbf{LIG} (Equation \eqref{gr}) and \textbf{Mixed Logit} (Theorem 3 in \cite{london2020bayesian}) policy classes. 

\begin{figure*}
     \centering
\includegraphics[width=.9439\textwidth]{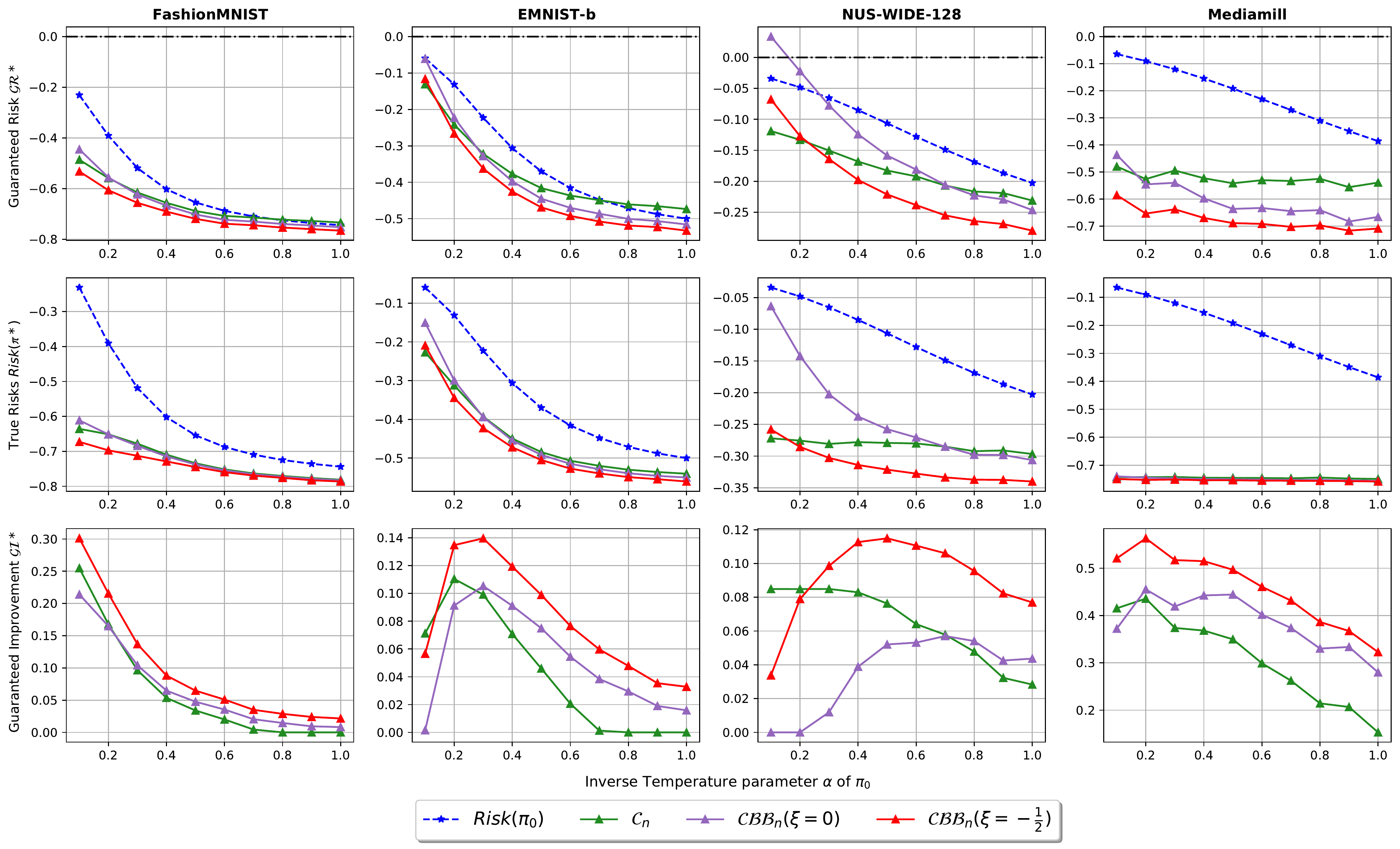}
    \caption{Behavior of the guaranteed risk $\mathcal{GR}^*$ ($\downarrow$ is better), the risk of the minimizer $\risk{\pi^*}$ ($\downarrow$ is better) and the guaranteed improvement $\mathcal{GI}^*$ ($\uparrow$ is better) given by the bounds (optimized with \textbf{LIG} policies) while changing $\pi_0$. We can observe that the newly proposed bounds can efficiently improve on $\pi_0$, with $\mcal{CBB}^\xi_n$ ($\xi = -\frac{1}{2}$) giving the best results.} 
    \label{fig:gi}
\end{figure*}

In Figure~\ref{fig:ls}, we plot $\risk{\pi_0}$, the true risk of $\pi_0$ alongside the best guaranteed risk $\mcal{GR}_{\mcal{LS}_n}^*$ given by the two policy classes,  while changing the logging policies $\pi_0$ (going from uniform to peaked policies by changing $\alpha$). 

We can observe, for the two policy classes and all scenarios considered, that this bound fails at providing  guaranteed improvement as its guaranteed risk $\mcal{GR}^*_{\mcal{LS}_n}$ is always smaller than $\risk{\pi_0}$ and is vacuous ($\mcal{GR}^*_{\mcal{LS}_n} \ge 0$) for some datasets. This bound is not tight enough to be used with our strategy.


\textbf{$\mcal{C}_n$ and $\mcal{CBB}^\xi_n$ do guarantee improvement.} Given logged data $\mcal{D}_n$, we optimize $\mcal{C}_n$ and $\mcal{CBB}^\xi_n$ for \textbf{LIG} policies (Equation \eqref{gr}) and plot in Figure \ref{fig:gi} the guaranteed risk $\mcal{GR}^*$, the true risk of the minimizer $\risk{\pi^*}$ as well as the positive guaranteed improvement by the bound $\max(\mcal{GI}^*, 0)$. 

To answer our question, we are interested in the first row (best guaranteed risk $\mcal{GR}^*$) and last row (best guaranteed improvement $\mcal{GI}^*$) of Figure \ref{fig:gi}. For the $\mcal{CBB}^\xi_n$ bound, we study particularly the two values of $\xi \in \{0, -\frac{1}{2}\}$.  

We can observe that contrary to $\mcal{LS}_n$, our bounds can guarantee improvement over $\pi_0$ in the majority of scenarios. 
 
 The $\mcal{C}_n$ bound gives great results when $\pi_0$ is close to uniform ($\alpha \rightarrow 0$) but fails sometimes (when $\alpha \rightarrow 1$) at improving the logging policy and one can observe that in the context of \textbf{FashionMNIST} and \textbf{EMNIST-b}.
 
 As for the $\mcal{CBB}_n^\xi$ bound, we can observe that choosing $\xi = -\frac{1}{2}$ consistently give the best results as it reduces considerably the dependency on $\mathcal{V}_{n}^\tau$. Note that $\mcal{CBB}^\xi$ with $\xi = -\frac{1}{2}$ never fails to produce guaranteed improvement across all settings. Having a uniform logging policy $\pi_0$ is the worst regime for this bound as $\mcal{V}^\tau_{n}$ reaches its highest value $1/\tau$. This is empirically confirmed with our experiments. We can see in Figure \ref{fig:gi} that, for small values of $\alpha$, the $\mcal{CBB}^\xi_n$ bound suffers the most; $\mcal{CBB}_n(\xi = 0)$ is always worse than $\mcal{C}_n$ and $\mcal{CBB}_n(\xi = -\frac{1}{2})$ produces worse guarantees than $\mcal{C}_n$ in the \textbf{NUS-WIDE-128} dataset. Once we drift away from uniform logging policies, the $\mcal{CBB}^\xi$ bound, especially with $\xi = - \frac{1}{2}$ gives the best guarantees on all the datasets considered.

\textbf{Tighter bounds give the best true risk.} In real-world problems, we cannot have access to $\risk{\pi^*}$ before deployment. In our experiments, we can compute $\risk{\pi^*}$ on the test sets as we have access to the true labels\footnote{up to a small $\mcal{O}(1/\sqrt{n_t})$ approximation error .}. We are interested in this quantity as we want to make sure that the bounds giving low guaranteed risk $\mcal{GR}^*$ will produce policies $\pi^*$ with low true risk $\risk{\pi^*}$. Even if the limiting behaviors of the bounds give an intuition of how they compare, this will further confirm that the gap between the bounds is not linked to constants but to quantities valuable to learning $\pi^*$. The second row on Figure \ref{fig:gi} confirms that the bounds with the best $\mcal{GR}^*$ reliably give the best $\risk{\pi^*}$ in all settings.

\textbf{Take away.} These experiments confirm that the policies $\pi^*$ obtained by optimising our newly proposed bounds improve, with high confidence, the logging policies $\pi_0$. The results also suggest the use of the variance sensitive bound $\mcal{CBB}_n(\xi = -1/2)$ for its consistent results across the different scenarios. However, if computing expectations under $\pi_0$ is difficult, one can adopt $\mcal{C}_n$ as it showed great results.

%
\section{Conclusion}
In this work, we introduce a new theoretically grounded strategy for offline policy optimization. This approach is based on generalization bounds, uses all the available data and does not require additional hyperparameters. Leveraging PAC-Bayesian tools, we provide novel generalization bounds tight enough to make our strategy viable, giving practitioners a principled way to confidently improve over the previous decision system offline. Our results can nicely be extended to learning efficient policies over slates \cite{slates} or continuous action policies \cite{continuous}. We believe that our work brings us closer to offline policy learning with online performance certificates. In the future, we would like to investigate tighter bounds for this problem and loosen our assumptions; e.g. to remove the need for having access to the logging policy $\pi_0$.

\section{Acknowledgments}

The authors are grateful for all the fruitful discussions engaged with David Rohde, Louis Faury and Imad Aouali which helped improve this work. We would also want to thank the reviewers for their valuable feedback.

\bibliography{bibliography}
\bibliographystyle{icml2023}

\newpage
\appendix
\onecolumn

\section{DISCUSSIONS AND PROOFS}

\subsection{Policies as mixtures of deterministic decision rules}
\label{mixpol}

As described in the paper, a policy $\pi$ takes a context $x\in\mcal{X}$ and defines a probability distribution over the $K$-dimensional simplex $\Delta_{K}$. In our work, we reinterpret policies as mixtures of deterministic decision rules. 

Let $f$ be the function that encodes the relevance of the action to the context $x$. Given a distribution $Q$ over the functions $f\in \mcal{F}_\Theta = \{f_\theta, \theta \in \Theta\}$, we define a policy as:
$$\forall x,a \quad \pi_Q(a|x) = \mathbbm{E}_{f \sim Q}\left[\mathbbm{1}\left[\operatorname*{argmax}_{a'} f(x,a') = a\right]\right].$$
A natural question is: can any policy $\pi$ be written in this form?

In general, the answer depends on the set $\mcal{F}_\Theta = \{f_\theta, \theta
\in \Theta\}$ we are considering. When the class $\mcal{F}_\Theta$ is rich enough, answer is yes, as proven by the following theorem.
\begin{theorem}
Let us fix a policy $\pi$. Let
$$\mathcal{G} = \{ g:\mathcal{X}\times\mathcal{A}\rightarrow \{0,1\} \text{ such that } \forall x, \exists ! a, g(x,a)=1 \}. $$
Then, there is a $\sigma$-algebra $\mathcal{S}$ on $\mathcal{G}$ and a probability distribution $Q_\pi$ on $(\mathcal{G},\mathcal{S})$ such that
$$\forall x,a \quad \pi(a|x) = \mathbbm{E}_{f \sim Q_\pi}\left[\mathbbm{1}\left[\operatorname*{argmax}_{a'} f(x,a') = a\right]\right].$$
\end{theorem}

\paragraph{Proof:} Fix a policy $\pi$.
Define the set $\Omega = [K]^\mathcal{X}$. That is, an element $\omega$ of $\Omega$ is a family of elements of $[K]$ indexed by $\mathcal{X}$: $\omega=(\omega_x)_{x\in\mathcal{X}}$.  Define the set of cylinders
$$ \mathcal{C} = \left\{ A \subset\Omega: A = \prod_{x\in\mathcal{X}}A_x \text{ and } {\rm card}(\{x:A_x\neq [K]\})<\infty \right\}.  $$
For such a set $A=\prod_{x\in\mathcal{X}}A_x$ we define
$$ P_\pi(A) = \prod_{x:A_x\neq \Omega}\left[ \sum_{a\in A_x} \pi(a|x) \right]. $$
Note in particular that, for a fixed $x\in\mathcal{X}$ and $a\in[K]$, we have
\begin{equation}
 \label{equa:p_pi}
 P_\pi(\{\omega\in\Omega: \omega_x = a \}) = \pi(a|x).
\end{equation}
Then, Kolmogorov extension theorem guarantees that there is a unique extension of $P_\pi$ to the $\sigma$-field $\mathcal{D}$ generated by $\mathcal{C}$, that is $\mathcal{D}=\sigma(\mathcal{C})$. We have thus built a probability space $(\Omega,\mathcal{D},P_\pi)$.

Now, for any $\omega=(\omega_x)_{x\in\mathcal{X}}$, we define the function $f_\omega:\mathcal{X}\times\mathcal{A}\rightarrow \{0,1\}$ by $f_\omega(x,a) = \mathbbm{1}\left[ \omega_x = a \right]$. Define, for any $C\in\mathcal{D}$, $S_C := \{f_\omega,\omega\in C\}$ and $Q_\pi(S_C) = P_\pi(C)$, and finally put $\mathcal{S}=\{S_C,C\in\mathcal{D}\}$. As the function $F:\omega\mapsto f_\omega$ is a bijection from $\Omega$ to $\mathcal{G}$, $\mathcal{S}$ is a $\sigma$-field and $Q_\pi$ is a probability distribution. We have thus equiped $\mathcal{G}$ with a $\sigma$-field $\mathcal{S}$ and a probability $Q_\pi$: $(\mathcal{G},\mathcal{S},Q_\pi)$ is a probability space.

Now, we check that
\begin{align*}
 \mathbbm{E}_{f \sim Q_\pi}\left[\mathbbm{1}\left[\operatorname*{argmax}_{a'} f(x,a') = a\right]\right]
 & =  Q_\pi\left(\left\{f\in\mathcal{G} : \operatorname*{argmax}_{a'} f(x,a') = a\right\}\right)
 \\
 & = P_\pi \left( \left\{ \omega\in\Omega: \operatorname*{argmax}_{a'} f_\omega(x,a') = a\right\} \right)
 \\
 & = P_\pi \left( \left\{ \omega\in\Omega: \omega_x = a \right\} \right)
 \\
 & = \pi(a|x)
\end{align*}
thanks to~\eqref{equa:p_pi}. This ends the proof.

\subsection{Proof of Proposition 1} \label{catoni_pbp}
Proposition \ref{PBP_catoni} is a direct application of \cite{CatoniBound}'s bound (see Theorem 3 in \cite{bguedj}) to the rescaled cIPS $0 \le 1 + \tau \cdot \cipsrisk{\cdot}{\tau} \le 1$ with deterministic decision functions $d_\theta$. Let us fix a prior $P$ over $\mcal{F}_\Theta$ and $\tau \in (0,1]$. For any $\delta \in (0,1]$, we have with probability at least $1 - \delta$ over draws of $\mcal{D}_n \sim (\nu, \pi_0)^n$: for any $Q$ that is $P$-continuous, any $\lambda >0$:
\begin{align*}
        1 + \tau \cdot \mathbbm{E}_{\theta \sim Q}\left[\mathbbm{E}_{(\nu, \pi_0)}\left[ \cipsrisk{d_\theta}{\tau} \right]\right] \le \frac{1}{(1 - e^{-\lambda})} \left( 1 - \exp \left[ - \lambda \cdot ( 1 + \tau \cdot \mathbbm{E}_{\theta \sim Q}[\cipsrisk{d_\theta}{\tau}]) - \frac{\mcal{KL}(Q||P) + \ln \frac{2\sqrt{n}}{\delta}}{n} \right] \right)
\end{align*}
with $KL[Q||P] = \mathbbm{E}_Q\left[ \ln Q/P \right]$. 

By linearity of the expectation and $\cipsrisk{\cdot}{\tau}$, we get:
\begin{align*}
        1 + \tau \cdot \mathbbm{E}_{(\nu, \pi_0)}\left[ \cipsrisk{\pi_Q}{\tau} \right] \le \frac{1}{(1 - e^{-\lambda})} \left( 1 - \exp \left[ - \lambda \cdot ( 1 + \tau \cdot \cipsrisk{\pi_Q}{\tau}) - \frac{KL[Q||P] + \ln \frac{2\sqrt{n}}{\delta}}{n} \right] \right).
\end{align*}
Rearranging the terms gives:
\begin{align*}
         \mathbbm{E}_{(\nu, \pi_0)}\left[ \cipsrisk{\pi_Q}{\tau} \right] &\le \frac{e^{-\lambda}}{\tau(1 - e^{-\lambda})} \left( 1 - \exp \left[ - \lambda \cdot \tau \cdot \cipsrisk{\pi_Q}{\tau} - \frac{KL[Q||P] + \ln \frac{2\sqrt{n}}{\delta}}{n} \right] \right)\\
         &\le \frac{1}{\tau(e^{\lambda} - 1)} \left( 1 - \exp \left[ - \lambda \cdot \tau \cdot \cipsrisk{\pi_Q}{\tau} - \frac{KL[Q||P] + \ln \frac{2\sqrt{n}}{\delta}}{n} \right] \right).
\end{align*}
The last step is to exploit the fact that the bias of $\cipsrisk{\cdot}{\tau}$ is negative (because the cost $c \le 0$), we have for any $\pi$:
\begin{align*}
    \mathbbm{E}_{x\sim\nu, a\sim\pi_0(\cdot|x)}\left[ \cipsrisk{\pi}{\tau} \right] &= \mathbbm{E}_{x\sim\nu, a\sim\pi_0(\cdot|x)}\left[\frac{\pi(a|x)}{\max(\pi_0(a|x), \tau)}c(x,a)\right]\\
    &\ge \mathbbm{E}_{x\sim\nu, a\sim\pi_0(\cdot|x)}\left[\frac{\pi(a|x)}{\pi_0(a|x)}c(x,a)\right] = \risk{\pi} 
\end{align*}
which gives the result stated in Proposition \ref{PBP_catoni} by taking the minimum over $\lambda > 0$:
\begin{align*}
    \risk{\pi_Q} \le \min_{\lambda > 0} \frac{1}{\tau(e^{\lambda} - 1)} \left( 1 - \exp \left[ - \lambda \cdot \tau \cdot \cipsrisk{\pi_Q}{\tau} - \frac{KL[Q||P] + \ln \frac{2\sqrt{n}}{\delta}}{n} \right] \right).
\end{align*}

\subsection{Limiting behavior of $\mcal{C}_n$} \label{catoni_lb}

To build an intuition of the dependency of $\mcal{C}_n$ on both $n$ and $\tau$, we can linearize the bound by exploiting the well-known inequality $1 - \exp(-x) \le x$ for $x \in [-1, 1]$. This gives:
$$\mcal{C}_n(\pi_Q) \le \min_{\lambda > 0} \frac{\lambda}{(e^{\lambda} - 1)} \cipsrisk{\pi_Q}{\tau} + \frac{KL[Q||P] + \ln \frac{2\sqrt{n}}{\delta}}{\tau(e^{\lambda} - 1) n}.$$
As the upper bound is a minimum over $\lambda > 0$, fixing $\lambda = \frac{1}{50}$ for example still gives a valid upper bound. This value leads to $\frac{\lambda}{(e^{\lambda} - 1)} \approx 1$ and $e^{\lambda} - 1 \approx \frac{1}{50}$, giving an approximated behaviour of the upper bound on $\mcal{C}_n$ of:
\begin{align*}
    \mcal{C}_n(\pi_Q) &\le  \cipsrisk{\pi_Q}{\tau} + 50 \cdot \frac{KL[Q||P] + \ln \frac{2\sqrt{n}}{\delta}}{\tau n} \\
    &\le  \cipsrisk{\pi_Q}{\tau} + \mcal{O}\left(\frac{KL[Q||P] + \ln \frac{2\sqrt{n}}{\delta}}{\tau n} \right).
\end{align*}

This result shows that $\mcal{C}_n$ improves the dependency on $n$ compared to $\mcal{LS}_n$.
\subsection{Proof of Theorem 1}
\label{thm1}

We fix $\mcal{D}_n \sim (\mu, \pi_0)^n, \tau \in (0, 1]$. To prove Theorem \ref{dominates}, we use the equality stated in Theorem 3 from \cite{bguedj} applied to the rescaled cIPS $0 \le \hat{\mcal{L}}_n(\cdot) = 1 + \tau \cdot \cipsrisk{\cdot}{\tau} \le 1$. For any distribution $P$, any distribution $Q$ that is $P$-continuous,  $\delta \in (0,1]$, we have:
\begin{align*}
    \sup_{0 \le p \le 1} \left\{p: kl(\hat{\mcal{L}}_n(\pi_Q)||p) \le \frac{KL[Q||P] + \ln \frac{2\sqrt{n}}{\delta}}{n} \right\} = 1 + \tau \cdot \mcal{C}^{P, \delta, \tau}_n(\pi_Q)
\end{align*}
with $\mcal{C}^{P, \delta, \tau}_n(\pi_Q) := \min\limits_{\lambda > 0} \dfrac{ 1 -
e^{ - \tau \lambda \Gamma_n^\tau(Q, \lambda, \delta)} }{\tau(e^{\lambda} -
1)}$, $\Gamma_n^\tau(Q, \lambda, \delta) = \cipsrisk{\pi_Q}{\tau} +
\frac{KL[Q||P] + \ln \frac{2\sqrt{n}}{\delta}}{\tau \lambda n}$ and $kl(q||p) =
q\log(\frac{q}{p}) + (1 - q)\log(\frac{1 - q}{1 - p})$, the KL divergence
between two Bernoulli variables of parameters $p$ and $q$. This means that:
\begin{align*}
    kl(\hat{\mcal{L}}_n(\pi_Q)||1 + \tau \cdot \mcal{C}^{P, \delta, \tau}_n(\pi_Q)) \le \frac{KL[Q||P] + \ln \frac{2\sqrt{n}}{\delta}}{n}.
\end{align*}
By leveraging the following inequality: $p \le q + \sqrt{2qkl(q||p)} +
2kl(q||p)$ for $p \leq q$, we get:
\begin{align*}
    1 + \tau \mcal{C}^{P, \delta, \tau}_n(\pi_Q) \le 1 + \tau \cipsrisk{\pi_Q}{\tau} + \sqrt{\frac{2[1 + \tau \cipsrisk{\pi_Q}{\tau}](KL[Q||P] + \ln \frac{2\sqrt{n}}{\delta})}{n}} + \frac{2(KL[Q||P] + \ln \frac{2\sqrt{n}}{\delta})}{n}.
\end{align*}
Giving the result of Theorem \ref{dominates}:
\begin{align*}
    \mcal{C}^{P, \delta, \tau}_n(\pi_Q) \le \cipsrisk{\pi_Q}{\tau} + \sqrt{\frac{2[\frac{1}{\tau} + \cipsrisk{\pi_Q}{\tau}](KL[Q||P] + \ln \frac{2\sqrt{n}}{\delta})}{\tau n}} + \frac{2(KL[Q||P] + \ln \frac{2\sqrt{n}}{\delta})}{\tau n}.
\end{align*}

\subsection{Proposition 3 beyond the i.i.d. case} \label{proof_prop3}

\begin{figure}[H]
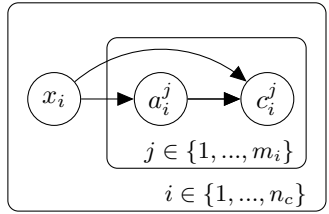

\centering
  \tikz{
 \node[latent] (x) {$x_i$};%
 \node[latent,right=of x,xshift=-0.3cm,fill] (a) {$a^j_i$}; %
 \node[latent,right=of a,xshift=-0.3cm] (c) {$c^j_i$}; %
\plate [inner sep=.25cm,yshift=.2cm, xshift = -.1cm] {plate1} {(a)(c)} {$j \in \{1,...,m_i$\}};%
 \plate [inner sep=.25cm,yshift=.2cm] {plate2} {(x) (plate1)} {$i \in \{1,...,n_c$\}};
 \edge {a} {c};
 \edge {a} {c};
 \edge {x} {a};
 \edge [out=40,in=140] {x} {c};
 }
 \caption{The "Multiple Interactions" Setting.}
 \label{dgp}
\end{figure}

In a multitude of applications, the i.i.d. assumption made on $\{x_i, a_i, c_i\}_{i\in[n]}$ can be violated. Indeed, a decision system can interact with the same context $x_i$ multiples times, trying different actions and logging the feedbacks as represented in Figure \ref{dgp}. Let $m_i$ be the number of times the system interacted with context $x_i$. The logged dataset in this case can be represented by 
$$\mcal{D}^{\{m_i\}_{i \in [n_c]}}_{n_c} = \left \{x_i, \{a_i^j, c_i^j\}_{j\in[m_i]}\right \}_{i\in[n_c]}$$
with $n_c$ representing the number of contexts and $n = \sum_i^{n_c} m_i$ the total number of datapoints. As soon as we have an $m_{i_0} > 1$, the i.i.d. assumption does not hold anymore as the samples $\{x_{i_0}, \{a_{i_0}^j, c_{i_0}^j\}_{j = 1}^{m_{i_0}}\}$ share the same observation $x_{i_0}$ and thus are dependent. In this case, the cvcIPS estimator will be written as:
$$\cvcipsrisk{\pi_Q}{\tau}{\xi} = \xi + \sum^{n_c}_{i = 1}\sum^{m_i}_{j = 1}\frac{\omega^\tau_{\pi_Q}(a^j_i|x_i)(c^j_i - \xi)}{n_c m_i}$$
We recover the i.i.d. case by taking $m_i = 1$ $\forall i$. Under this weaker assumption, \cite{CatoniBound} or any classical PAC-Bayesian bound cannot be applied directly.

\subsubsection{Proof of
Proposition 3}

In this section, we begin by stating Proposition \ref{pbp_tract_bernstein_cv}
for the more general case where we have a logged dataset $\mcal{D}^{\{m_i\}_{i
\in [n_c]}}_{n_c}$.

\begin{proposition}\label{bern_gen} Given a prior $P$ on $\mcal{F}_\Theta$, $\xi \in [-1, 0], \tau \in (0, 1]$, $\delta \in (0,1]$ and a set of strictly positive scalars $\Lambda = \{ \lambda_i \}_{i \in [n_\Lambda]}$. We have with probability at least $1 - \delta$ over draws of $\mcal{D}^{\{m_i\}_{i \in [n_c]}}_{n_c} \sim \prod_{i = 1}^{n_c}(\nu, \pi_0^{m_i})$: For any $Q$ that is $P$-continuous, any $\lambda \in \Lambda$:
\vspace{-1em}

\begin{align*}
  \risk{\pi_Q} &\le \cvcipsrisk{\pi_Q}{\tau}{\xi} - \xi \mathcal{B}^{\tau}_{n_c}(\pi_Q ) + \sqrt{\frac{KL[Q||P] + \ln \frac{4\sqrt{n_c}}{\delta}}{2 n_c}} \\
  &+ \frac{KL[Q||P] + \ln \frac{2 n_\lambda}{\delta}}{\lambda} + \frac{\lambda l_\xi}{n_c} \sum_{i = 1}^{n_c}\frac{1}{m_i n_c} g\left(\frac{\lambda b_\xi}{m_i n_c} \right) \mathcal{V}^{\tau, i}(\pi_Q)
\end{align*}

with $g : u \xrightarrow[]{}\frac{\exp(u) - 1 - u}{u^2}$, $l_\xi = \max \left(\xi^2, (1 + \xi)^2 \right)$, $b_\xi = \frac{1 + \xi}{\tau} - \xi$, $\mathcal{V}^{\tau, i}(\pi) = \mathbbm{E}_{\pi(.|x_i)} \left[\frac{\pi_0(a|x_i)}{\max(\tau, \pi_0(a|x_i))^2}\right]$ and \\
$\mcal{B}^{\tau}_{n_c}(\pi) = \frac{1}{n_c}\sum_{i=1}^{n_c} \mathbbm{E}_{\pi(.|x_i)}\left[\mathbbm{1}[\pi_0(a|x_i) < \tau]  \left( 1 - \frac{\pi_0(a|x_i)}{\tau} \right)\right]$.
\end{proposition}

We use a decomposition similar to \cite{snips-claire} and rewrite the difference $\risk{\pi_Q} - \cvcipsrisk{\pi_Q}{\tau}{\xi}= D_1(\pi_Q) + D_2(\pi_Q) + D_3(\pi_Q)$ with:
\begin{align*}
D_1(\pi_Q) &= \risk{\pi_Q} - \frac{1}{n_c}\sum_{i = 1}^{n_c}\risk{\pi_Q|x_i} \\
D_2(\pi_Q) &= \frac{1}{n_c}\sum_{i = 1}^{n_c}\risk{\pi_Q|x_i} - \frac{1}{n_c}\sum_{i = 1}^{n_c} \xi + \mathbbm{E}_{a \sim \pi_0(\cdot|x_i)}\left[\omega^\tau_{\pi_Q}(a^j_i|x_i)(c(a, x_i) - \xi) \right] \\
D_3(\pi_Q) &= \frac{1}{n_c}\sum_{i = 1}^{n_c} \xi + \mathbbm{E}_{\pi_0(\cdot|x_i)}\left[\omega^\tau_{\pi_Q}(a|x_i)(c(a, x_i) - \xi) \right] - \cvcipsrisk{\pi_Q}{\tau}{\xi}.
\end{align*}

For the first difference $D_1$, we use \cite{mcallester} bound for the $[0, 1]$-bounded loss using $0 \le 1 + \risk{\pi_Q|x_i} \le 1$. We get with probability at least $1 - \delta$, For any $Q$ that is $P$-continuous:
\begin{align}
\label{mc1}
    D_1(\pi_Q) \le \sqrt{\frac{KL[Q||P] + \ln \frac{2\sqrt{n_c}}{\delta}}{2n_c}}.
\end{align}

The second difference quantifies the bias of our estimator given the contexts
$\{x_i, ..., x_{n_c}\}$. Even if we cannot compute it, we can give an upper bound for $D_2$. We have:
\begin{align*}
    D_2(\pi_Q) &= \frac{1}{n_c}\sum_{i = 1}^{n_c}\risk{\pi_Q|x_i} - \frac{1}{n_c}\sum_{i = 1}^{n_c} \xi + \mathbbm{E}_{a \sim \pi_0(\cdot|x_i)}\left[\omega^\tau_{\pi_Q}(a|x_i)(c(a, x_i) - \xi) \right]\\
    &= \frac{1}{n_c}\sum_{i = 1}^{n_c} \mathbbm{E}_{a \sim \pi_0(\cdot|x_i)}\left[(\omega^0_{\pi_Q}(a|x_i) - \omega^\tau_{\pi_Q}(a|x_i))(c(a, x_i) - \xi) \right]\\
    &= \frac{1}{n_c}\sum_{i = 1}^{n_c} \mathbbm{E}_{a \sim \pi_0(\cdot|x_i)}\left[\mathbbm{1}[\pi_0(a|x_i) < \tau](\frac{\pi_Q(a|x_i)}{\pi_0(a|x_i)} - \frac{\pi_Q(a|x_i)}{\tau})(c(a, x_i) - \xi) \right]\\
    &= \frac{1}{n_c}\sum_{i = 1}^{n_c} \mathbbm{E}_{a \sim \pi_Q(\cdot|x_i)}\left[\mathbbm{1}[\pi_0(a|x_i) < \tau](1 - \frac{\pi_0(a|x_i)}{\tau})(c(a, x_i) - \xi) \right] (c \le 0)\\
    &\le - \frac{\xi}{n_c}\sum_{i = 1}^{n_c} \mathbbm{E}_{a \sim \pi_Q(\cdot|x_i)}\left[\mathbbm{1}[\pi_0(a|x_i) < \tau](1 - \frac{\pi_0(a|x_i)}{\tau}) \right] = - \xi \mcal{B}^{\tau}_{n_c}(\pi_Q).
\end{align*}

We obtain $- \xi \mcal{B}^{\tau}_{n_c}(\pi_Q)$, an empirical upper bound to $D_2(\pi_Q)$.

The last step is to control the difference $D_3$. Before doing this, we need to state two lemmas that will help us control the difference $D_3$.

\begin{lemma}{Change of measure:} \label{pac_bayes}
    Let $f$ be a function of the parameter $\theta$ and data $S$, for any distribution Q that is P continuous, for any $\delta \in (0,1]$, we have with probability $1 - \delta$ :
    \begin{align}
        \mathbbm{E}_{\theta \sim Q}[f(\theta,S)] \le KL[Q||P] + \ln \frac{\Psi_f}{\delta} 
    \end{align}
    with $\Psi_f = \mathbbm{E}_{S}\mathbbm{E}_{\theta \sim P}[e^{f(\theta,S)}]$.
\end{lemma}
Lemma \ref{pac_bayes} is the backbone of many PAC Bayes bounds. It is proven in many references, see for example~\cite{bayes_alquier} or Lemma 1.1.3 in~\cite{CatoniBound}.
We will combine it with an inequality on the moment generating function to prove a Bernstein-like PAC-Bayes bound \cite{seldin}.

\begin{lemma}{} \label{expo_bound}
    Let $W$ be a r.v with $\mathbbm{E}[W^2] < \infty$, we suppose that $\mathbbm{E}[W] - W \le B$. Let $g : u \xrightarrow[]{}\frac{\exp(u) - 1 - u}{u^2}$, we have for all $\eta \ge 0$:
    \begin{align}
        \mathbbm{E}[\exp(\eta (\mathbbm{E}[W] - W) - \eta^2 g(\eta B)\mathbbm{V}[W])] \le 1.
    \end{align}
\end{lemma}
Lemma~\ref{expo_bound} is stated and proven in~\cite{mcdiarmid1998concentration}.

Combining both lemmas allows us to control the difference $D_3$ with a conditional Bernstein PAC-Bayesian bound:

\begin{corollary}{Conditional Bernstein PAC-Bayesian Bound:}
    Let's fix a $\lambda > 0$ and a prior $P$, for any distribution $Q$ that is $P$ continuous, for any $\delta \in (0,1]$, we have with probability at least $1 - \delta$:
\begin{align}
    D_3(\pi_Q) &\le \frac{KL[Q||P] + \ln \frac{1}{\delta}}{\lambda} + \frac{\lambda l_\xi}{n_c} \sum_{i = 1}^{n_c}\frac{1}{m_i n_c} g\left(\frac{\lambda b_\xi}{m_i n_c} \right) \mathcal{V}^{\tau, i}(\pi_Q)
\end{align}
with $l_\xi = \max \left(\xi^2, (1 + \xi)^2 \right)$, $b_\xi = \frac{1 + \xi}{\tau} - \xi$, $\mathcal{V}^{\tau, i}(\pi) = \mathbbm{E}_{\pi(.|x_i)} \left[\frac{\pi_0(a|x_i)}{\max(\tau, \pi_0(a|x_i))^2}\right]$.
\end{corollary}
\paragraph{Proof:} Let us fix a context $x_i$ and an action $a_i^j$ and let $\theta \sim P$. We have:
$$D^j_i(\theta) = \mathbbm{E}_{\pi_0(\cdot|x_i)}\left[\omega^\tau_{d_\theta}(a|x_i)(c(a, x_i) - \xi) \right] - \omega^\tau_{d_\theta}(a^j_i|x_i)(c(a_i^j, x_i) - \xi) \le b_\xi = \frac{1 + \xi}{\tau} - \xi .$$
We fix a $\lambda$ and choose: 
\begin{align*}
    f(\theta,S) &= \sum_{i = 1}^{n_c}  \sum_{j = 1}^{m_i} \left[ \frac{\lambda}{m_i n_c} D^j_i(\theta) - (\frac{\lambda}{m_i n_c})^2 g\left(\frac{\lambda b_\xi}{m_i n_c} \right) \mathbbm{E}_{\pi_0(\cdot|x_i)}[D_i(\theta)^2]\right]\\
    &= \sum_{i = 1}^{n_c}  \sum_{j = 1}^{m_i} \left[ \Delta_i^j(\theta) \right].
\end{align*}

From Lemma 2 and because the prior $P$ does not depend on the data, we have:
\begin{align*}
\Psi_f = \mathbbm{E}_{\prod_i \pi_0(\cdot|x_i)}\mathbbm{E}_{\theta \sim P}[e^{f(\theta,S)}] &= \mathbbm{E}_{\theta \sim P}\mathbbm{E}_{\prod_i \pi_0(\cdot|x_i)}[e^{f(\theta,S)}]\\
&= \mathbbm{E}_{\theta \sim P} \prod_i (\mathbbm{E}_{\pi_0(\cdot|x_i)}[e^{\Delta_i^0(\theta)}])^{m_i} \le 1.
\end{align*}

It means that $\ln \Psi_f \le 0$. Using this in Lemma 1, we get:

\begin{align*}
    D_3(\pi_Q) &\le \frac{KL[Q||P] + \ln \frac{1}{\delta}}{\lambda} + \sum_{i = 1}^{n_c}  \sum_{j = 1}^{m_i}\frac{\lambda}{(m_i n_c)^2} g\left(\frac{\lambda b_\xi}{m_i n_c} \right) \mathbbm{E}_{\pi_0(\cdot|x_i)}\left[\mathbbm{E}_{\theta \sim Q}[D_i(\theta)^2]\right] \\
    &= \frac{KL[Q||P] + \ln \frac{1}{\delta}}{\lambda} + \frac{\lambda}{n_c} \sum_{i = 1}^{n_c}\frac{1}{m_i n_c} g\left(\frac{\lambda b_\xi}{m_i n_c} \right) \mathbbm{E}_{\theta \sim Q}\left[\mathbbm{E}_{\pi_0(\cdot|x_i)}\left[D_i(\theta)^2\right]\right].
\end{align*}

we also use the following inequality to upper bound $\mathbbm{E}_{\pi_0(\cdot|x_i)}[D_i(\theta)^2]$:
\begin{align*}
    \mathbbm{E}_{\pi_0(\cdot|x_i)}[D_i(\theta)^2] &\le \mathbbm{E}_{a \sim \pi_0(\cdot|x_i)}\left[\frac{d_\theta(a|x_i)}{\max(\pi_0(a|x_i), \tau)^2}(c(a, x_i) - \xi)^2 \right] \\
    &\le \max(\xi^2, (1 + \xi)^2) \mathbbm{E}_{a \sim \pi_0(\cdot|x_i)}\left[\frac{d_\theta(a|x_i)}{\max(\pi_0(a|x_i), \tau)^2} \right] \quad \text{because both  } c, \xi \in [-1, 0]\\
    &= l_\xi \mathcal{V}^{\tau, i}(d_\theta).
\end{align*}

As the quantity $\mathcal{V}^{\tau, i}$ is linear in $d_\theta$, the result in Corollary 1 follows:

\begin{align*}
    D_3(\pi_Q) &\le \frac{KL[Q||P] + \ln \frac{1}{\delta}}{\lambda} + \frac{\lambda l_\xi}{n_c} \sum_{i = 1}^{n_c}\frac{1}{m_i n_c} g\left(\frac{\lambda b_\xi}{m_i n_c} \right) \mathcal{V}^{\tau, i}(\pi_Q).
\end{align*}

Finally, We take a union bound of Corollary 1 over $\Lambda$, a discrete set with cardinal $n_\Lambda$, and combine its result with the bound giving~\eqref{mc1} through another union bound to obtain Proposition \ref{bern_gen}.

\subsubsection{Choice of $\Lambda$ when $m_i = m$}
When the number of interactions $m$ is constant across all contexts, the result in Corollary 1 becomes for a fixed $\lambda$:
\begin{align*}
    D_3(\pi_Q) &\le \frac{KL[Q||P] + \ln \frac{1}{\delta}}{\lambda n} + \lambda l_\xi g\left(\lambda b_\xi \right) \mathcal{V}^{\tau}_{n_c}(\pi_Q)
\end{align*}
where $\mathcal{V}^\tau_{n_c}(\pi_Q)$ was defined in Proposition \ref{pbp_tract_bernstein_cv}. 

We would like to choose a $\lambda$ that minimizes the bound on
$D_3$. Unfortunately, we cannot do it because the minimizer $\lambda^*$ depends on $Q$. Instead, we build an interval in which $\lambda^*$ can be found. 

The function $g : u \xrightarrow[]{}\frac{\exp(u) - 1 - u}{u^2}$ behaves like $\frac{\exp(u)}{u^2}$ when $u$ is big enough, meaning that we should control the values of $g$, and thus $\lambda$ by an upper bound. Choosing $\lambda \le b = \frac{2n}{b_\xi}$ allows us to control the function $g\left(\frac{\lambda b_\xi}{n} \right) \le g(2) \le 1.1$. 

Now that an upper bound is found, we still need to find the lowest possible value for $\lambda^*$. Of course, choosing the interval $[0, b]$ can be enough but we want to do more than that. $\lambda^*$ verifies the following equality:

\begin{align*}
    \lambda^* &= \sqrt{\frac{KL[Q||P] + \ln \frac{1}{\delta}}{\frac{l_\xi}{n}g\left(\frac{\lambda^* b_\xi}{n} \right) \mathcal{V}^{\tau}_{n_c}(\pi_Q) + \frac{\lambda^* l_\xi b_\xi}{n^2}g'\left(\frac{\lambda^* b_\xi}{n} \right) \mathcal{V}^{\tau}_{n_c}(\pi_Q)}}.
\end{align*}

Let's assume that $\lambda^\star \leq b$. (If not, we can still restrict to
$\lambda\in[a, b]$, with the value of $a$ found below.)
We have that $KL[Q||P] \ge 0$, and
$\mathcal{V}^{\tau}_{n_c} \le
\frac{1}{\tau}$. As the function $g$ is increasing and convex ($g'$
increasing), we get the following inequality:

\begin{align*}
    \lambda^* \ge \sqrt{\frac{n \tau \ln\frac{1}{\delta}}{l_\xi g(2) + 2 l_\xi g'(2)}}.
\end{align*}

Using the fact that  $g'(2) = 1/2$ and $g(2) + 1 \le 5/2$, we get: 
$$\lambda^* \ge \sqrt{\frac{n \tau \ln\frac{1}{\delta}}{l_\xi g(2) + l_\xi}} \ge \sqrt{\frac{2 n \tau \ln\frac{1}{\delta}}{5 l_\xi}} = a.$$

We now have an interval $\lambda^* \in [a, b]$. One can observe that the optimal $\mcal{O}(\sqrt{n}) \le \lambda^* \le \mcal{O}(n)$.

We choose the set $\Lambda$ to be a linear discretization of $[a, b]$ giving $\Lambda = \{a + i(b - a)\}_{i \in [n_\Lambda]}$.

\subsubsection{Dependencies of the bound}
\label{bigo}
The bound for the i.i.d. case can be written as:
\begin{align*}
  \risk{\pi_Q} &\le \cvcipsrisk{\pi_Q}{\tau}{\xi} - \xi \mathcal{B}^{\tau}_{n}(\pi_Q) + \sqrt{\frac{\mcal{KL}(Q||P) + \ln \frac{4\sqrt{n}}{\delta}}{2 n}} + \min\limits_{\lambda \in \Lambda} \left\{\frac{\mcal{KL}(Q||P) + \ln \frac{2 n_\Lambda}{\delta}}{\lambda n} +  \lambda l_\xi g\left(\lambda b_\xi \right) \mathcal{V}_{n}^\tau(\pi_Q)\right\}
\end{align*}

with $\Lambda = \{\frac{a}{b} + i \frac{(b - a)}{n}\}_{i \in [n_\Lambda]}$. We know that the biggest value of $\lambda \in \Lambda$ is $b = \frac{2}{b_\xi}$ and that $g(2) \approx 1$. This gives:
\begin{align*}
    \min\limits_{\lambda \in \Lambda} A(\lambda) &= \min\limits_{\lambda \in \Lambda} \frac{\mcal{KL}(Q||P) + \ln \frac{2 n_\Lambda}{\delta}}{\lambda n} +  \lambda l_\xi g\left(\lambda b_\xi \right) \mathcal{V}_{n}^\tau(\pi_Q)\\
    &\le \min\limits_{\lambda \in \Lambda}\frac{\mcal{KL}(Q||P) + \ln \frac{2 n_\Lambda}{\delta}}{\lambda n} +  \lambda l_\xi g\left(2 \right) \mathcal{V}_{n}^\tau(\pi_Q) \\
    &\lessapprox \min\limits_{\lambda \in \mathbbm{R}^+}\frac{\mcal{KL}(Q||P) + \ln \frac{2 n_\Lambda}{\delta}}{\lambda n} +  \lambda l_\xi \mathcal{V}_{n}^\tau(\pi_Q) = 2 \sqrt{\frac{\l_\xi \mathcal{V}_{n}^\tau(\pi_Q) \left(\mcal{KL}(Q||P) + \ln \frac{2 n_\Lambda}{\delta} \right)}{n}}
\end{align*}

We make the hypothesis that our $\Lambda$ is well built to have a value of $\lambda$ close to the true minimizer most of the time. This gives the following limiting behavior:
\begin{align*}
    \mcal{CBB}^{\xi}_{n}(\pi_Q) &= \cvcipsrisk{\pi_Q}{\tau}{\xi} - \xi \mathcal{B}^{\tau}_{n}(\pi_Q) + \mathcal{O}\left( \left(\frac{1}{2\sqrt{2}} + \sqrt{l_\xi \mathcal{V}_{n}^\tau(\pi_Q)}\right)\sqrt{ \frac{\mathcal{KL}(Q||P)}{n}}\right).
\end{align*}

\subsection{Linear Independent Gaussian Policies}
\label{ligp}
To obtain these policies, we restrict $f_\theta$ to:
\begin{align}
    \forall x,a \quad f_\theta(x, a) = \phi(x)^T\theta_a
\end{align}
with $\phi$ a fixed transform over the contexts. This results in a parameter $\theta$ of dimension $d = p \times K$ with $p$ the dimension of the features $\phi(x)$ and $K$ the number of actions. We also restrict the family of distributions $\mcal{Q}_{d + 1} = \{Q_{\boldsymbol{\mu}, \sigma} = \mathcal{N}(\boldsymbol{\mu}, \sigma^2 I_d), \boldsymbol{\mu} \in \mathbbm{R}^d, \sigma > 0\}$ to independent Gaussians with shared scale. 

Estimating the propensity of $a$ given $x$ reduces the computation to a one dimensional integral:
\begin{align*}
    \pi_{\boldsymbol{\mu}, \sigma}(a|x) &= \mathbbm{E}_{\epsilon \sim \mathcal{N}(0, 1)}\left[\prod_{a' \neq a} \Phi\left(\epsilon + \frac{\phi(x)^T(\boldsymbol{\mu}_a - \boldsymbol{\mu}_{a'})}{\sigma ||\phi(x)||}\right) \right]
\end{align*}
with $\Phi$ the cumulative distribution function of the standard normal.
\paragraph{Proof:} We rewrite the definition of $\pi_{\boldsymbol{\mu}, \sigma}$ as a probability and exploit the stability of the Gaussian distribution.
\begin{align*}
    \pi_{\boldsymbol{\mu}, \sigma}(a|x) &= \mathbbm{E}_{\theta \sim \mathcal{N}(\boldsymbol{\mu}, \sigma^2 I_d)}\left[ \mathbbm{1}[\operatorname{argmax}_{a'}\phi(x)^T \theta_{a'} = a] \right]\\
    &= \mathbbm{E}_{S \sim \mathcal{N}(\phi(x)^T\boldsymbol{\mu}, \sigma^2 ||\phi(x)||^2 I_K)}\left[ \mathbbm{1}[\operatorname{argmax}_{a'} S_{a'} = a] \right] \\
    &= \mathbbm{P}_{S \sim \mathcal{N}(\phi(x)^T\boldsymbol{\mu}, \sigma^2 ||\phi(x)||^2 I_K)}\left(\operatorname{argmax}_{a'} S_{a'} = a \right) \\
    &= \mathbbm{P}_{S \sim \mathcal{N}(\phi(x)^T\boldsymbol{\mu}, \sigma^2||\phi(x)||^2 I_K)}\left(S_{a} \ge S_{a'}, \quad \forall a' \neq a \right)\\
    &= \mathbbm{P}_{Z \sim \mathcal{N}(0_K, I_K)}\left(Z_{a} + \frac{\phi(x)^T(\boldsymbol{\mu}_a - \boldsymbol{\mu}_{a'})}{\sigma ||\phi(x)||} \ge  Z_{a'}, \quad \forall a' \neq a \right).
\end{align*}
We condition on $Z_a$ to obtain independent events as for all $a$, the random variables $Z_a$ are independent.
\begin{align*}
    \pi_{\boldsymbol{\mu}, \sigma}(a|x) &= \mathbbm{P}_{Z \sim \mathcal{N}(0_K, I_K)}\left(Z_{a} + \frac{\phi(x)^T(\boldsymbol{\mu}_a - \boldsymbol{\mu}_{a'})}{\sigma ||\phi(x)||} \ge  Z_{a'}, \quad \forall a' \neq a \right) \\
    &= \mathbbm{E}_{\epsilon \sim \mathcal{N}(0, 1)} \left[\mathbbm{P}_{Z \sim \mathcal{N}(0_K, I_K)}\left(\epsilon + \frac{\phi(x)^T(\boldsymbol{\mu}_a - \boldsymbol{\mu}_{a'})}{\sigma ||\phi(x)||} \ge  Z_{a'}, \quad \forall a' \neq a| Z_a = \epsilon \right) \right] \\
    &= \mathbbm{E}_{\epsilon \sim \mathcal{N}(0, 1)}\left[ \prod_{a' \neq a} \mathbbm{P}_{z \sim \mathcal{N}(0, 1)}\left(z \le \epsilon + \frac{\phi(x)^T(\boldsymbol{\mu}_a - \boldsymbol{\mu}_{a'})}{\sigma ||\phi(x)||} \right) \right]\\
    &= \mathbbm{E}_{\epsilon \sim \mathcal{N}(0, 1)}\left[\prod_{a' \neq a} \Phi\left(\epsilon + \frac{\phi(x)^T(\boldsymbol{\mu}_a - \boldsymbol{\mu}_{a'})}{\sigma ||\phi(x)||}\right) \right] \qed.
\end{align*}

\subsection{Why not Mixed Logit Policies?} 
\label{mixedl}
\cite{london2020bayesian} used in their analysis Mixed Logit Policies to derive a learning principle for softmax policies.  Mixed Logit Policies can be written as:
$$\forall (a,x), \quad \pi^{ML}_{\boldsymbol{\mu}, \sigma}(a|x) = \mathbbm{E}_{\theta \sim \mathcal{N}(\boldsymbol{\mu}, \sigma^2 I_d)}[\operatorname{softmax}_K(\phi(x)^T\theta_a)].$$

Even if these policies can behave properly (reparametrization trick gradient for instance), they are not ideal for learning with guarantees in the context of Offline Contextual Bandits. Indeed, we know that the solution of the contextual bandit problem is a deterministic decision function $d^*$, always choosing the action with the minimum cost. Let us suppose that there exists a parameter $\mu^*$ such that: 
$$\forall (a,x), \quad d^*(a|x) = d_{\mu^*}(a|x) = \mathbbm{1}[\operatorname{argmax}_{a' \in \mcal{A}}(\phi(x)^T\mu^*_{a'}) = a]$$
We also suppose that we have access to its parameter $\mu^*$. To recover $d_{\mu^*}$ with \textbf{LIG} policies, we need to have the scale parameter small enough $\sigma \rightarrow 0$ as :
$$  \pi_{\boldsymbol{\mu^*}, \sigma}(a|x) \xrightarrow[\sigma \rightarrow 0]{} d_{\mu^*}(a|x) \quad \forall x,a .$$

For \textbf{Mixed Logit} policies however, having $\sigma \rightarrow 0$ is not enough as:
$$  \pi^{ML}_{\boldsymbol{\mu^*}, \sigma}(a|x) \xrightarrow[\sigma \rightarrow 0]{} \operatorname{softmax}_K(\phi(x)^T\mu^*_a) \quad \forall x,a .$$

One should also increase the norm of $\mu^*$ enough ($||\mu^*|| \rightarrow \infty$) to obtain $d_{\mu^*}$.

Let us suppose that we start with the same prior $P = \mcal{N}(\boldsymbol{\mu^*}, I_d)$ in our bounds. The price to pay in terms of complexity $KL(Q_{\boldsymbol{\mu}, \sigma}||P)$ to obtain the solution; a deterministic policy, will be much higher for \textbf{Mixed Logit} policies (as we should decrease $\sigma$ and increase the norm of $\boldsymbol{\mu}$) than \textbf{LIG} policies (only decrease $\sigma$ and let $\boldsymbol{\mu} = \boldsymbol{\mu^*}$). This means that for a fixed number of samples $n$, we will always get better results with \textbf{LIG} policies than \textbf{Mixed Logit} policies.

\subsection{The bounds stated for LIG policies} \label{lig_bounds}

In this section, we want to state the previous Propositions \ref{PBP_catoni} and \ref{pbp_tract_bernstein_cv} (valid for any policy) for the class of \textbf{LIG} policies. This class of policies uses Independent Gaussian distributions with shared scale so we will begin by stating the KL divergence between $P = \mcal{N}(\boldsymbol{\mu}_0, \sigma_0 I_d)$ and $Q = \mcal{N}(\boldsymbol{\mu}, \sigma I_d)$. We have:
\begin{align*}
    KL[Q||P] = D[\boldsymbol{\mu}, \sigma, \boldsymbol{\mu_0}, \sigma_0] = \frac{|| \mu - \mu_0||^2}{2\sigma_0^2} + d \left(\frac{\sigma^2}{2\sigma_0^2} + \ln \frac{\sigma_0}{\sigma} - \frac{1}{2} \right).
\end{align*}

We write the bounds slightly differently by taking the minimimum over the considered $\lambda$ (if the bound is true for any $\lambda$, it is true for the minimum of the bound over $\lambda$).

We state Catoni's bound for \textbf{LIG} policies:

\begin{corollary}{\textbf{LIG} policies with Catoni's bound}  

Given a Gaussian prior $P = \mcal{N}(\boldsymbol{\mu}_0, \sigma_0 I_d)$, $\tau \in (0,1]$, $\delta \in (0,1]$. We have with probability $1 - \delta$ over draws of $\mcal{D}_n \sim (\nu, \pi_0)^n$: 

$\forall \boldsymbol{\mu} \in \mathbbm{R}^d, \sigma > 0$:
\begin{align*}
        \risk{\pi_{\boldsymbol{\mu}, \sigma}} \le \min \limits_{\lambda > 0} \frac{1}{\tau(e^{\lambda} - 1)} \left[ 1 - \exp\left( -  \tau \lambda \cipsrisk{\pi_{\boldsymbol{\mu}, \sigma}}{\tau} + \frac{ D[\boldsymbol{\mu}, \sigma, \boldsymbol{\mu_0}, \sigma_0] + \ln \frac{2\sqrt{n}}{\delta}}{n} \right) \right]
\end{align*}
\end{corollary}

We call $\mcal{C}_n(\pi_{\boldsymbol{\mu}, \sigma})$ the upper bound stated by this corollary. We get:
\begin{align*}
\mcal{GR}^*_{\mcal{C}} &= \min_{\pi_{\boldsymbol{\mu}, \sigma}} \mcal{C}_n(\pi_{\boldsymbol{\mu}, \sigma})\\
\pi^*_{\mcal{C}} &= \argmin_{\pi_{\boldsymbol{\mu}, \sigma}} \mcal{C}_n(\pi_{\boldsymbol{\mu}, \sigma})\\
\mcal{GI}^*_{\mcal{C}} &= \risk{\pi_0} - \mcal{GR}^*_{\mcal{C}}.
\end{align*}

Similarly, we state our variance sensitive bound for \textbf{LIG} policies:

\begin{corollary}{\textbf{LIG} policies variance sensitive bound.} 

Given a Gaussian prior $P = \mcal{N}(\boldsymbol{\mu}_0, \sigma_0 I_d)$, $\xi \in [-1, 0], \tau \in (0, 1]$, $\delta \in (0,1]$ and a set of strictly positive scalars $\Lambda = \{ \lambda_i \}_{i \in [n_\Lambda]}$. We have with probability at least $1 - \delta$ over draws of $\mcal{D}^m_{n_c} \sim \prod_{i = 1}^{n_c}(\nu, \pi_0^m)$: 

$\forall \boldsymbol{\mu} \in \mathbbm{R}^d, \sigma > 0$:
\vspace{-3em}

\begin{align*}
  \risk{\pi_{\boldsymbol{\mu}, \sigma}} &\le \cvcipsrisk{\pi_{\boldsymbol{\mu}, \sigma}}{\tau}{\xi} - \xi \mathcal{B}^{\tau}_{n_c}(\pi_{\boldsymbol{\mu}, \sigma}) + \sqrt{\frac{D[\boldsymbol{\mu}, \sigma, \boldsymbol{\mu_0}, \sigma_0] + \ln \frac{4\sqrt{n_c}}{\delta}}{2 n_c}} \\
  & + \min \limits_{\lambda \in \Lambda} \left \{\frac{D[\boldsymbol{\mu}, \sigma, \boldsymbol{\mu_0}, \sigma_0] + \ln \frac{2 n_\lambda}{\delta}}{\lambda} + \frac{\lambda l_\xi}{n} g\left(\frac{\lambda b_\xi}{n} \right) \mathcal{V}_{n_c}^\tau(\pi_{\boldsymbol{\mu}, \sigma}) \right \}
\end{align*}
\end{corollary}

We call $\mcal{CBB}_n(\pi_{\boldsymbol{\mu}, \sigma}, \xi, m)$ the upper bound stated by this corollary. Similarly we get:
\begin{align*}
\mcal{GR}^*_{\mcal{CBB}(\xi, m)} &= \min_{\pi_{\boldsymbol{\mu}, \sigma}} \mcal{CBB}_n(\pi_{\boldsymbol{\mu}, \sigma}, \xi, m)\\
\pi^*_{\mcal{CBB}(\xi, m)} &= \argmin_{\pi_{\boldsymbol{\mu}, \sigma}} \mcal{CBB}_n(\pi_{\boldsymbol{\mu}, \sigma}, \xi, m)\\
\mcal{GI}^*_{\mcal{CBB}(\xi, m)} &= \risk{\pi_0} - \mcal{GR}^*_{\mcal{CBB}(\xi, m)}.
\end{align*}

\section{EXPERIMENTS}

\subsection{Detailed Statistics of the dataset splits used}
\label{detailed_exps}

As described in the experiments section, we use the supervised to bandit conversion to simulate logged data as previously adopted in the majority of the literature \cite{swaminathan2015batch, snips, london2020bayesian, faury20distributionally, dro_2}. In this procedure, you need a split $D_l$ (of size $n_l$) to train the logging policy $\pi_0$, another split $D_c$ (of size $n_c$) to generate the logging feedback with $\pi_0$, and finally a test split $D_{test}$ (of size $n_{test}$) to compute the true risk $\risk{\pi}$ of any policy $\pi$. In our experiments, we split the training split $D_{train}$ (of size $N$) of the four datasets considered into $D_l$ ($n_l = 0.05N$) and $D_c$ ($n_c = 0.95N$) and use their test split $D_{test}$. The detailed statistics of the different splits can be found in Table \ref{table:det_stats}.

\begin{table}[H]
\centering
\begin{tabular}{ |c||c|c|c|c|c|c|}
 \hline
 Datasets & $N$ & $n_l$ & $n_c$ & $n_{test}$ & $K$ & $p$ \\
 \hline
 \textbf{FashionMNIST} &  60 000 &  3000 &  57 000 &  10 000  & 10 & 784\\
 \textbf{EMNIST-b} & 112 800 &  5640 &  107 160 &  18 800  & 47 & 784\\
 \textbf{NUS-WIDE-128} & 161 789 &  8089 &   153 700 &  107 859  & 81 & 128\\
  \textbf{Mediamill} &   30 993 &  1549 &  29 444 & 12 914 & 101 & 120\\
 \hline
\end{tabular}
\caption{Detailed statistics of the splits used.}
\label{table:det_stats}
\end{table}

\subsection{Detailed hyperparameters}
\label{training}

Contrary to previous work, our method does not require tuning any loss function hyperparameter over a hold out set. We do however need to choose parameters to optimize the policies. 

\paragraph{The logging policy $\pi_0$.} $\pi_0$ is trained on $D_l$ (supervised manner) with the following parameters:
\begin{itemize}
    \item We use $L_2$ regularization of $10^{-6}$. This is used to prevent the logging policy $\pi_0$ from being close to deterministic, allowing efficient learning with importance sampling.
    \item We use Adam \cite{adam} with a learning rate of $10^{-1}$ for $10$ epochs.
\end{itemize}

\paragraph{Optimising the bounds.} All the bounds are optimized with the following parameters:
\begin{itemize}
    \item The clipping parameter $\tau$ is fixed to $1/K$ with $K$ the action size of the dataset.
    \item We use Adam \cite{adam} with a learning rate of $10^{-3}$ for $100$ epochs.
    \item For the bounds optimized over \textbf{LIG} policies, the gradient is a one dimensional integral, and is approximated using $S = 32$ samples.

\begin{align*}
    \pi_{\boldsymbol{\mu}, \sigma}(a|x) &= \mathbbm{E}_{\epsilon \sim \mathcal{N}(0, 1)}\left[\prod_{a' \neq a} \Phi\left(\epsilon + \frac{\phi(x)^T(\boldsymbol{\mu}_a - \boldsymbol{\mu}_{a'})}{\sigma ||\phi(x)||}\right) \right] \\
    &\approx \frac{1}{S} \sum_{s = 1}^S \prod_{a' \neq a} \Phi\left(\epsilon_s + \frac{\phi(x)^T(\boldsymbol{\mu}_a - \boldsymbol{\mu}_{a'})}{\sigma ||\phi(x)||}\right) \quad \epsilon_1, ..., \epsilon_S \sim \mcal{N}(0, 1).
    \end{align*}

    \item For $\mcal{C}_n$, we treat $\lambda$ as a parameter and we look for the minimum of the bound with respect to $\mu, \sigma$ and $\lambda$.
    \item For $\mcal{CBB}^\xi_n$, we choose the size of $\Lambda$ to be $n_\Lambda = 100$ and for each iteration $j$ of the optimization procedure, we take $\lambda_j \in \Lambda$ that minimizes the estimated bound and proceed to compute the gradient w.r.t $\mu$ and $\sigma$ with $\lambda_j$. 

\end{itemize}

\subsection{Impact of changing the number of interactions $m$}

\begin{figure*}
     \centering
\includegraphics[width=\textwidth]{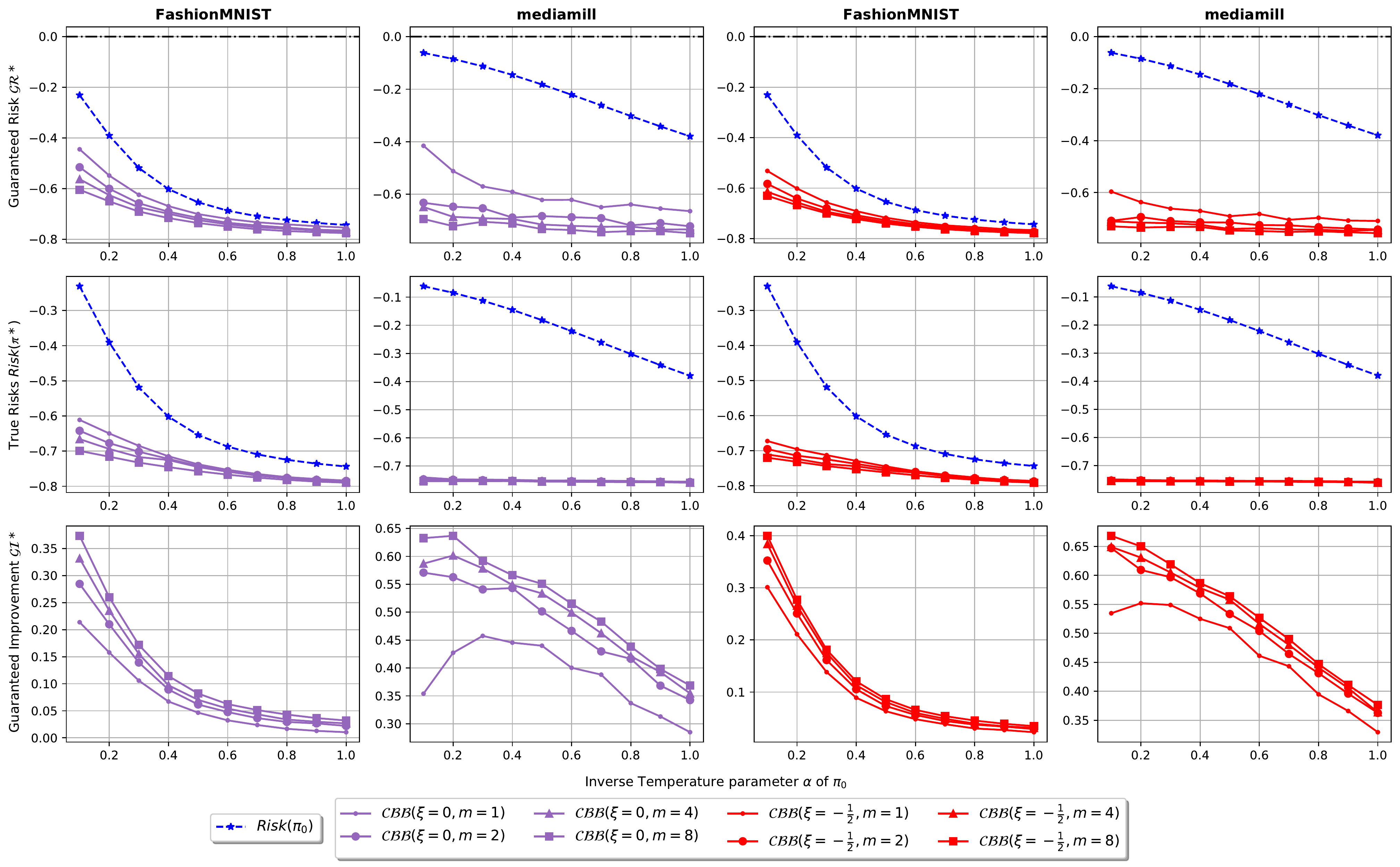}
    \caption{Behavior of the guaranteed risk $\mathcal{GR}^*$ ($\downarrow$ is better), the risk of the minimizer $\risk{\pi^*}$ ($\downarrow$ is better) and the guaranteed improvement $\mathcal{GI}^*$ ($\uparrow$ is better) given by changing the number of interactions $m$ and $\pi_0$.}
    \label{fig:supp}
\end{figure*}

The bound proposed in Proposition \ref{pbp_tract_bernstein_cv} can work beyond the i.i.d. setting and applies to the "multiple interactions" case. Intuitively, adding more interactions with the contexts $x$ allows us to reduce the uncertainty on the cost and thus learn better policies. We want to explore this in Figure \ref{fig:supp}. We construct with $\pi_0$ a logged dataset with the number of interactions $m \in \{1, 2, 4, 8 \}$ using both \textbf{FashionMNIST} and \textbf{Mediamill} datasets. Once $m > 1$, we can only use the $\mcal{CBB}$ bound. We stick to the values of $\xi$ previously used $\xi \in \{0, -1/2\}$.

We can observe that increasing the number of $m$ consistently give better results, in terms of guarantees and also the quality of the policy $\pi^*$ minimizing the bounds. We can also observe that even though $m$ reduces the gap between the two estimators ($\xi = 0$ compared to $\xi = -1/2$), the cvcIPS estimator with $\xi = -1/2$ still gives the best results.

\end{document}